\documentclass[10pt,twocolumn,letterpaper]{article}
\usepackage[pagenumbers]{cvpr}

\usepackage{fontawesome5}
\usepackage{tabularx}
\usepackage{lipsum}
\usepackage[table,xcdraw]{xcolor}
\usepackage{hhline}
\usepackage{bm}
\usepackage{bookmark}
\usepackage{fancyhdr}
\usepackage{subcaption}
\usepackage{wrapfig}
\usepackage{stmaryrd}
\usepackage{rotating}
\usepackage{mwe}
\usepackage{multicol}
\usepackage{multirow}
\usepackage{tikz}
\usepackage{environ}
\usepackage{adjustbox}
\usepackage{siunitx}
\usepackage{pifont}
\usepackage{mathtools}
\usepackage{amsmath,amssymb}
\usepackage{tcolorbox}
\usepackage{mdframed}
\usepackage{booktabs}
\usepackage{overpic}
\usepackage[usestackEOL]{stackengine}
\usepackage[normalem]{ulem}

\definecolor{cvprblue}{rgb}{0.21,0.49,0.74}
\definecolor{TableGray1}{HTML}{9B9B9B}
\definecolor{TableGray2}{HTML}{C0C0C0}
\definecolor{TableGray3}{HTML}{EFEFEF}

\newcommand{\myparagraph}[1]{\noindent \textbf{#1} --}
\newcommand{\myparagraphno}[1]{\noindent \textbf{#1}}

\definecolor{coolred}{HTML}{E77475}
\definecolor{coolblue}{HTML}{277C9D}
\definecolor{coolgreen}{HTML}{598938}
\definecolor{coolyellow}{HTML}{FACB77}
\definecolor{coolorange}{HTML}{FF8C00}
\definecolor{coolcyan}{HTML}{3EC3B2}
\definecolor{coollightblue}{HTML}{62BCDD}
\definecolor{coolpurple}{HTML}{976AA3}

\definecolor{agentours}{HTML}{FAE0D0}
\definecolor{agentinstant}{HTML}{D8E8F4}
\definecolor{agentd4}{HTML}{FFF1C7}

\definecolor{probinglinear}{HTML}{F9A868}
\definecolor{probinglinearpact}{HTML}{5AB75F}
\definecolor{probinggru}{HTML}{61A1D2}

\def\cA{{\mathcal{A}}}

\def\cC{{\mathcal{C}}}
\def\cD{{\mathcal{D}}}

\def\cP{{\mathcal{P}}}

\def\cT{{\mathcal{T}}}

\def\bma{{\bm{a}}}

\def\bmg{{\bm{g}}}
\def\bmh{{\bm{h}}}

\def\bmo{{\bm{o}}}
\def\bmp{{\bm{p}}}

\newcommand{\yes}{\ding{51}}
\newcommand{\no}{\ding{55}}

\newcolumntype{Y}{>{\centering\arraybackslash}p}
\newcolumntype{Z}{>{\centering\arraybackslash}X}
\newcolumntype{C}{>{\columncolor{TableGray2}}c}
\newcolumntype{L}{>{\columncolor{TableGray2}}l}
\newcolumntype{P}{>{\centering\arraybackslash\columncolor{blue!20}}X}
\newcolumntype{N}{>{\centering\arraybackslash\columncolor{orange!20}}X}
\newcolumntype{R}{>{\centering\arraybackslash\columncolor{green!20}}X}

\newcommand{\epbox}[1]{\tcbox[on line,colframe=black,boxsep=0pt,left=1pt,right=1pt,top=0.4pt,bottom=0.4pt,boxrule=0.4pt]{#1}}

\newcommand{\env}{${\scriptstyle\sim}$}

\newcommand{\office}{\epbox{Test-bldg/20}}
\newcommand{\officesub}{\epbox{Test-bldg/14}}

\newcommand{\realbox}[1]{\tcbox[on line,colframe=white,boxsep=0pt,left=1pt,right=1pt,top=0pt,bottom=0pt,colback=green!20]{#1}}
\newcommand{\simbox}[1]{\tcbox[on line,colframe=white,boxsep=0pt,left=1pt,right=1pt,top=0pt,bottom=0pt,colback=orange!20]{#1}}
\newcommand{\trainbox}[1]{\tcbox[on line,colframe=white,boxsep=0pt,left=1pt,right=1pt,top=0pt,bottom=0pt,colback=blue!20]{#1}}

\newcommand{\dfourbox}[1]{\tcbox[on line,colframe=white,boxsep=0pt,left=1pt,right=1pt,top=0pt,bottom=0pt,colback=agentd4]{#1}}
\newcommand{\instantbox}[1]{\tcbox[on line,colframe=white,boxsep=0pt,left=1pt,right=1pt,top=0pt,bottom=0pt,colback=agentinstant]{#1}}
\newcommand{\oursbox}[1]{\tcbox[on line,colframe=white,boxsep=0pt,left=1pt,right=1pt,top=0pt,bottom=0pt,colback=agentours]{#1}}

\newcommand{\reallabel}[1]{#1}

\newcommand{\rem}[1]{\texttt{\scriptsize {// #1}}}

\newcommand{\suppm}[1]{sup. mat.~\ref{#1}}

\newcommand{\urlwebsitehttp}{https://europe.naverlabs.com/research/publications/reasoning-in-visual-navigation-of-end-to-end-trained-agents}

\newcommand{\mysection}[2]{\setcounter{section}{#1}\section{#2}}

\newcommand\extrafootertext[1]{%
    \bgroup
    \renewcommand\thefootnote{\fnsymbol{footnote}}%
    \renewcommand\thempfootnote{\fnsymbol{mpfootnote}}%
    \footnotetext[0]{\hspace{-1em}#1}%
    \egroup
}

\def\confName{CVPR}
\def\confYear{2025}
\title{Reasoning in visual navigation of end-to-end trained agents: a dynamical systems approach}

\author{
Steeven Janny$^1$
\and
Hervé Poirier$^1$
\and
Leonid Antsfeld$^1$
\and
Guillaume Bono$^1$
\and
Gianluca Monaci$^1$
\and
Boris Chidlovskii$^1$
\and
Francesco Giuliari$^3$
\and
Alessio Del Bue$^2$
\and
Christian Wolf$^1$
}

\def\numepisodes{262}

\begin{document}
\maketitle

\newcommand{\revision}[1]{%
  \iftoggle{cvprfinal}{%
    #1 %
  }{%
    \textcolor{red}{#1} %
  }%
}

\begin{abstract}
Progress in Embodied AI has made it possible for end-to-end-trained agents to navigate in photo-realistic environments with high-level reasoning and zero-shot or language-conditioned behavior, but benchmarks are still dominated by simulation. In this work, we focus on the fine-grained behavior of fast-moving real robots and present a large-scale experimental study involving \numepisodes{} navigation episodes in a real environment with a physical robot, where we analyze the type of reasoning emerging from end-to-end training. In particular, we study the presence of realistic dynamics which the agent learned for open-loop forecasting, and their interplay with sensing. We analyze the way the agent uses latent memory to hold elements of the scene structure and information gathered during exploration. We probe the planning capabilities of the agent, and find in its memory evidence for somewhat precise plans over a limited horizon. Furthermore, we show in a post-hoc analysis that the value function learned by the agent relates to long-term planning. Put together, our experiments paint a new picture on how using tools from computer vision and sequential decision making have led to new capabilities in robotics and control. An interactive tool is available \href{\urlwebsitehttp}{[here]}.
\end{abstract}

\vspace{-4mm}
\section{Introduction}
\extrafootertext{$^1$ Naver Labs Europe, France. {\scriptsize \texttt{firstname.lastname@naverlabs.com}}}
\extrafootertext{$^2$ Istituto Italiano di Tecnologia, Genova, Italy.}
\extrafootertext{$^3$ Fondazione Bruno Kessler, Trento, Italy.}

\begin{figure}[t] \centering
    \includegraphics[width=0.9\linewidth]{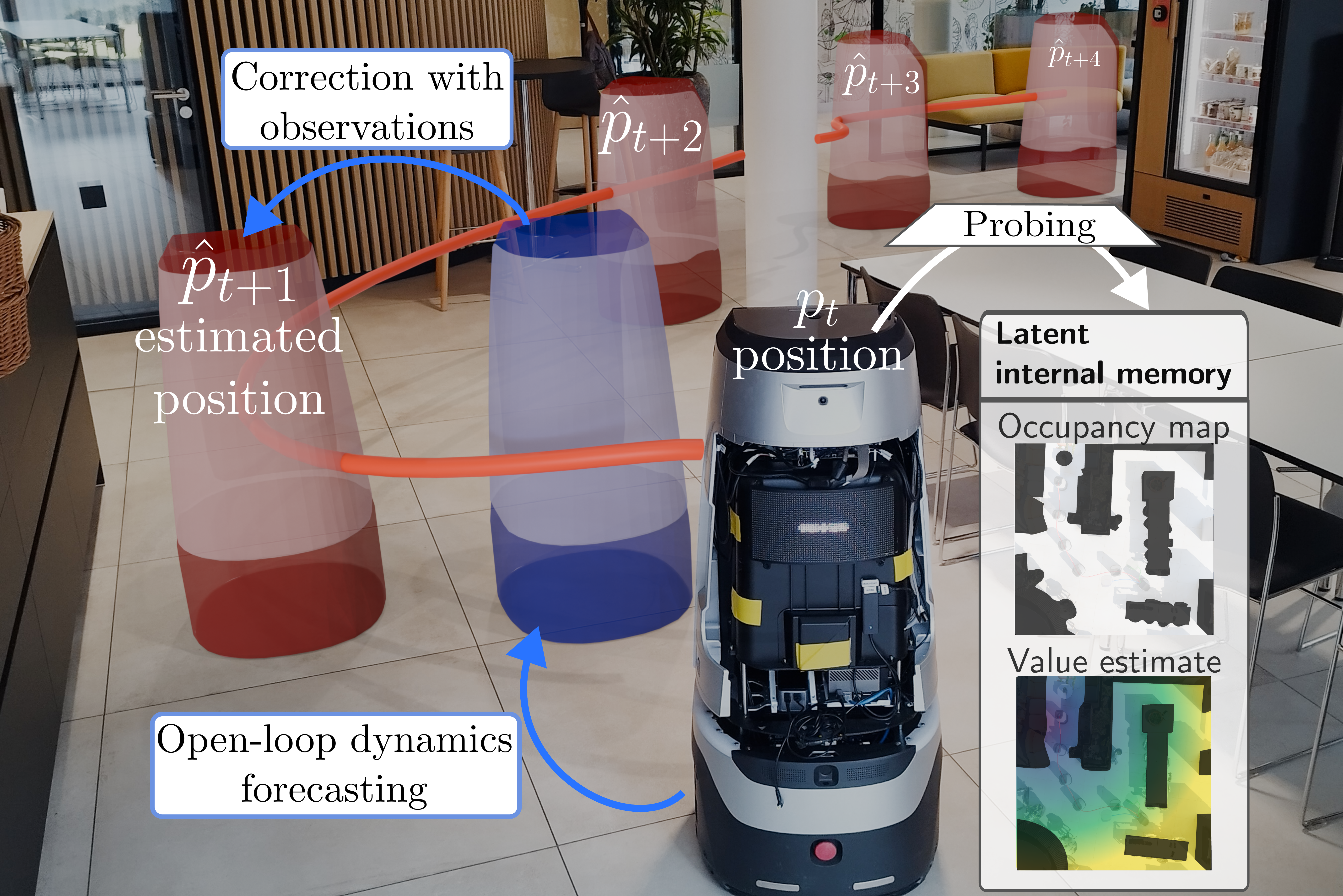}
    \caption{\label{fig:teaser}In a large-scale analysis of \numepisodes{} episodes of a real robot in a real environment, we report on the type of reasoning emerging after end-to-end training agents with realistic motion: they learn a dynamical motion model exploited with open-loop forecasting and corrected by sensing, latent scene structure, exploration information, and long-term value estimates.}   
    \vspace*{-3mm}
\end{figure}
Historically, the importance of computer vision for robot navigation tasks largely depended on which field addressed the problem. On one hand, roboticists consider navigation as a perception and reconstruction problem combined with planning. This naturally required robots to deal with noise and uncertainty \cite{TBF2002probabilisticrobotics} and thus computer vision, leading to the development of visual odometry and SLAM solutions \cite{TBF2002probabilisticrobotics,murORB2,LSDSLAM2014,imap2021,gsslam2024}. On the other hand, research in machine learning addressed it as a form of sequential decision making and a large body of work modeled it as a POMDP solved with Reinforcement Learning (RL). Early work was vision-less and focused on simple grid-like environments, finite states and observations and tabular representations. The introduction of function approximation into RL enabled end-to-end training of agents in photo-realistic environments like Habitat \cite{Savva_2019_ICCV} and AI2Thor \cite{ai2thor}, and very recently on large-scale offline datasets like Open-X-Embodiment \cite{openxembodiment2024}.
Now, computer vision is enabling a switch to realistic conditions in what is now known as Embodied AI (EAI).

End-to-end navigation policies are typically trained with RL in an environment with simple motion --- stepwise teleportation, leading to learning high-level decisions only. When deployed on a real robot, the displacements predicted by the policy are passed to a classical low-level controller, tasked with reaching the desired position. While this strategy led to breakthroughs in navigation, it results in very slow motion, as the agent stops between each step, and heavily relies on the quality of the low-level controller.

Interestingly, robotic manipulation followed a similar track, but moved rapidly to dynamic-aware environments \cite{chua2018deep,andrychowicz2020learning,loquercio2021learning}, integrating a dynamical model to mimic the behavior of a real robot, which has been shown to be instrumental for sim2real transfer. Recent work \cite{bono2024learning} indicates that this also applies to navigation: empowering the simulator with a numerical model of the robot motion plays an important role in transferring the navigation skills of the end-to-end agent to a real robot. This policy does not perform position control, but outputs target velocities, which are input to a dynamical model simulating the behavior of the real physical robot, enabling smoothly and quick navigation.

In this work we raise the question on what is actually learned by these policies. In contrast to a large part of the EAI community, which recently focused on tasks requiring high-level reasoning and semantics, up to language understanding, we here aim to \textit{understand} the capabilities of trained agents in another task dimension: acting for precise motion enabled by fine-grained geometry/perception and reliable estimates of robot dynamics. There is evidence, for instance, that generalist robot policies are capable of addressing broad classes of tasks, but tend to fail at skills which require finer precision in motion planning \cite{nakamoto2024steering}. An essential question remains: \textit{what are the inner workings of agents trained on large-scale visual data?}

We experimentally show that a policy trained end-to-end with RL leads to the internal identification of a latent dynamical model from its interaction with the simulator alone, if realistic motion is simulated. We show that prediction/correction steps akin to a Kalman filter are learned, introduce a method to estimate the importance of each step by analyzing sensitivity of the policy to changes of the real dynamics of the agent, and show that a potential gap in dynamics can be closed with proper techniques. We introduce a new probing method evaluating the dynamics prediction performance of an agent.

We further show how the agent memory encodes information on the scene structure and on the explored area, and we measure the impact of memory ablations. We probe the role of long-horizon planning in these agents through comparisons with experts planners and by analyzing the agent's value function. 
All together, these experiments, a majority of which have been executed with real robots in a real environment, shed light on how large-scale training on sequential visual data with realistic motion leads to the emergence of a useful dynamical model in visual agents.

\section{Related work}
\label{sec:related_work}

\myparagraphno{Visual navigation} has been classically solved in robotics using explicit modeling~\citep{burgard1998interactive,macenski2020marathon,marder2010office}, which requires solutions for mapping and localization~\citep{bresson2017simultaneous, labbe19rtabmap,thrun2005probabilistic}, 
for planning~\citep{konolige2000gradient, sethian1996fast} and for low-level control \citep{fox1997dynamic,rosmann2015timed}. These methods depend on accurate sensor models, filtering, dynamical models and optimization. End-to-end trained models directly map input to actions and are typically trained with RL~\citep{DBLP:conf/iclr/JaderbergMCSLSK17,mirowski17learning,zeng2024poliformer,uppal2024spin} or IL~\citep{DBLP:conf/nips/DingFAP19}. They learn flat representations~\cite{bono2024learning}, occupancy maps~\citep{Chaplot2020Learning}, semantic maps~\citep{chaplot2020object}, latent metric maps~\citep{DBLP:conf/pkdd/BeechingD0020,Henriques_2018_CVPR,DBLP:conf/iclr/ParisottoS18}, topological maps~\citep{BeechingECCV2020,Chaplot_2020_CVPR,shah2022viking}, explicit episodic memory~\citep{chen_think_2022,du2021vtnet,Fang_2019_CVPR,reed_generalist_2022}, implicit representations~\citep{Marza2022NERF} or navigability~\cite{Mole2024}.
Our study targets end-to-end policies featuring recurrent memory and benefiting from an additional motion model in simulation.
\myparagraphno{Dynamical models} were rarely considered as key concern for learning-based solutions in indoor navigation, which typically focus on planning and high-level control in simulation \citep{chaplot2020object,Chaplot2020Learning,gervet2023navigating,CrocoNav2024} and rely on handcrafted low-level controllers for deployment. The notion of dynamics is more common in manipulation where joint inertia, response time and slack must be simulated to accurately reproduce displacement and interactions \cite{chua2018deep,andrychowicz2020learning,loquercio2021learning}; consequently, they played a large role in sim2real gap mitigation and domain adaptation \cite{van2019sim,eysenbach2020off}. 
Frequent choices are domain randomization and a stochastic dynamical model to maintain good behavior on real platform \cite{chebotar2019closing,muratore2022robot,peng2018sim,mankowitz2020robust}.  
In navigation, the separation between high-level and low-level controller, as in \citep{chaplot2020object,Chaplot2020Learning,gervet2023navigating,CrocoNav2024}, leads to delays and limits reactivity. Our work builds on \cite{bono2024learning} which solves this problem by augmenting the Habitat simulator with a dedicated realistic motion.
\myparagraph{Learning dynamical systems} A large body of work also focuses explicitly on learning dynamical systems from offline datasets of \{action, observations\} pairs, a task inherited from \textit{System Identification} in control theory. Learned models then target different tasks, such as simulation \citep{pfaff2020learning,bauersfeld2021neurobem,janny2024space}, state estimation \citep{stubberud1991neural,de2007nonlinear} or control \citep{bauersfeld2021neurobem,zoboli2023deep,janny2021deep} by coupling the learned model with a classical algorithm.
\myparagraph{Model-based ML} Learning accurate models of the environment has been classically done in RL as \textit{State Representation Learning} \cite{lesort_state_2018}. Model-based RL directly integrates a model of the environment into the learning algorithm and/or the agent, eg. as \textit{World Models} \cite{Ha2018WorldModels,hafner2022dreamer}. They have seen a recent resurgence in variants trained large-scale with next-token prediction \cite{hu2023gaia1,zhang2024copilotd}. Our work is based on model-free RL, which obtains SOTA performance in the targeted navigation tasks, and evaluates what kind of models are learned from large-scale training.
\myparagraph{Evaluating navigation models} 
The sim2real gap represents one of the main challenges \cite{hofer2020sim2real,kadian_sim2real_2020,peng2018sim}.
Experiments with mobile robots are time-consuming, 
and the reproduction of identical evaluation conditions is difficult \cite{SadekICRA2022}. Evaluation in calibrated simulation has been proposed as an alternative \cite{kadian_sim2real_2020}.

\begin{figure*}
\begin{minipage}{0.25\textwidth}
        \begin{tikzpicture}
            \draw (0, 0) node[anchor=west, inner sep=0] {\includegraphics[width=\linewidth]{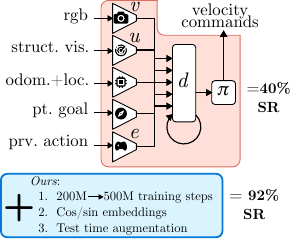}};        
            \draw (2.4, -0.55) node[anchor=west, inner sep=0] {\tiny Bono  et al. \cite{bono2024learning}};
        \end{tikzpicture}
        \caption{ \revision{\label{fig:bono_et_al}We build upon \cite{bono2024learning} with improvements allowing the SR in real setup to increase by +50\%.}}
\end{minipage}
\hfill
\begin{minipage}{0.73\textwidth}
    \small 
    \centering
    \setlength{\tabcolsep}{1mm}
    \begin{tabularx}{\linewidth}{lcNNNNNNRRR}
        \specialrule{1pt}{0pt}{0pt}
        \rowcolor{TableGray2}
        \textbf{Method} &  &
        \multicolumn{3}{c}{\cellcolor{orange!60} \shortstack{Sim (+dyn.) \\\epbox{HM3D/2.5k}}} &
        \multicolumn{3}{c}{\cellcolor{orange!60} \shortstack{Sim (+dyn.)\\ \office}} &
        \multicolumn{3}{c}{\cellcolor{green!50}\shortstack{Real\\ \office}} 
        \\
        \rowcolor{TableGray2}
        \textbf{\& action space} & 
            \multirow{-2}{*}{
                 \begin{sideways}
                    \textbf{ dyn*}
                \end{sideways}
            } &
        {\cellcolor{orange!60}\bf  SR} & {\cellcolor{orange!60}\bf  SPL} & {\cellcolor{orange!60}\bf SCT} &
        {\cellcolor{orange!60}\bf  SR} & {\cellcolor{orange!60}\bf  SPL} & {\cellcolor{orange!60}\bf SCT} &
        {\cellcolor{green!50}\bf  SR} & {\cellcolor{green!50}\bf  SPL} & {\cellcolor{green!50}\bf SCT} 
        \\ \specialrule{1pt}{0pt}{0pt}
        \cellcolor{agentd4}
        \textbf{(a) D4} \cite{sadek2022indepth} & 
        \cellcolor{agentd4} \no &
        {29.1} & {18.1} & {2.0} &
        {10.0} & {7.3} & {0.9} &
        ~~0.0~~~~~~ & ~~0.0~~~~~~ & ~~0.0~~~~~~ 
        \\
        \cellcolor{agentinstant}
        \textbf{(b) D28-instant} \cite{yokoyama2021success} & 
        \cellcolor{agentinstant} \no &
        27.6 & 11.6 & 5.0 &
        25.0 & 11.1 & 5.8 &
        10.0~~~~~~ & ~~5.3~~~~~~ & ~~1.7~~~~~~
        \\
        \cellcolor{agentours}
        \textbf{(c) D28-dynamics} \cite{bono2024learning} & 
        \cellcolor{agentours} \yes &
        {\bf 97.6} & {\bf 82.3} & {\bf 52.5} &
        {\bf 100.0} & {\bf 82.1} & {\bf 64.5} &
        {\bf 92.5} & {\bf 61.1} & {\bf 22.4}  
        \\
        \cellcolor{agentours}&\cellcolor{agentours}&&&&&&&
        {\bf \scriptsize$\pm$2.9} & {\bf \scriptsize$\pm$5.0} & {\bf \scriptsize$\pm$3.0}  
        \\ \specialrule{1pt}{0pt}{0pt}        
    \end{tabularx}
    
    \vspace*{-3mm}
    \caption{\label{tab:variants}Comparison of agents: 
    \dfourbox{(a)} 4 motion commands $\{${\footnotesize \texttt{FORWARD 25cm}, \texttt{TURN\_LEFT $10^{\circ}$}, \texttt{TURN\_RIGHT $10^{\circ}$}, \texttt{STOP}}$\}$, no dynamical model; \instantbox{(b)} 28 pairs of instant+constant velocities (no dynamical model); \oursbox{(c)} 28 pairs of velocities+identified realistic dynamical model. This agent has been evaluated in the \reallabel{real scenario} 4 times with 20 episodes each, we report mean and std.dev over these 4 experiments. *training only.}
\end{minipage}
\vspace{-5mm}
\end{figure*}

\section{\revision{Preliminaries}}
\label{sec:agent}

We \revision{reproduced} the work by Bono et al. \cite{bono2024learning} and use PPO to train a policy in simulation. The conclusions of this work should hold for a variety of navigation tasks, but for the sake of simplicity we focus on point goal navigation of fast moving agents in real environments with a particular emphasis on sim2real transfer. \revision{In this section, we recall their setup and state our modifications, which are not considered as contributions. More details are in \suppm{sec:model-details}}. The agent receives a set of observations at each time step $t$ in the form of RGB images $\mathbf{I}_t$ and a Lidar-like vector of ranges $\mathbf{S}_t {\in}\mathbb{R}^{K}$, ``scan'' and takes actions $\mathbf{a}_t$ to reach a goal position given as polar coordinates relative to its starting position.
In our case, the scan $\mathbf{S}_t$ is collected by a set of four \textit{RealSense} depth sensors%
. The action space is discrete, each action class $a$ among the $28$ possibilities corresponds to a pair $(a_v, a_\omega)$ of linear and angular velocity commands. %

The agent does not have direct access to a map and outputs actions from sensory inputs. However, some form of onboard localization wrt. the starting position at $t{=}0$ is required, which allows the task to communicate the goal to the agent. We provide the agent with two forms of localization: 
integrated odometry sensed from wheel encoders $\hat{\mathbf{p}}^r_t$, and \textit{Adaptive Monte Carlo Localization} from the ROS package AMCL, which uses a 1D-Lidar, denoted as $\hat{\mathbf{p}}^a_t$. Both are given wrt. the start of the episode, as are the static point goal coordinates $\mathbf{g}_0$. Both inputs contains estimated current position and velocity.
Similar to a large body of work \cite{SDMIA15-Hausknecht,DBLP:conf/iclr/JaderbergMCSLSK17,khanna2024goatbench,bongratz2024chooseRLAlg,CrocoNav2024}, the agent disposes of a latent memory vector $\mathbf{h}_t$, defined as the hidden state of a recurrent network,
\setlength{\abovedisplayskip}{3pt}
\setlength{\belowdisplayskip}{3pt}
\begin{align}
    \mathbf{h}_t = & d(\mathbf{h}_{t-1}, v(\mathbf{I}_t), u(\mathbf{S}_t), \mathbf{g}_0, \hat{\mathbf{p}}^r_t, \hat{\mathbf{p}}^a_t, e(\mathbf{a}_{t-1})),
    \label{eq:gru}
    \\
    \mathbf{a}_t = & \pi(\mathbf{h}_t), 
    \label{eq:policy}
\end{align}
where $d$ is a two-layer GRU with hidden state $\mathbf{h}_t$, and gating equations have been omitted for ease of notation.
The different inputs go through dedicated encoders: a ResNet-18 $v(\cdot)$, a 1D-CNN $u(\cdot)$ and an embedding $e$ of the previous action. The other inputs are encoded with MLPs, omitted from the notation. The policy $\pi$ is linear. 

The goal $\mathbf{g}_0$ is static, ie. constant over the episode and given in the coordinate frame centered on the agent at $t{=}0$. As the agent needs to transform it into its egocentric frame while moving, we add an auxiliary linear head $l$ predicting dynamically the relative goal position $\hat{\mathbf{g}}_t = l(\mathbf{h}_{t})$, supervised from privileged information in simulation. 

\myparagraph{Training for realistic motion} As in \cite{bono2024learning}, we integrate realistic motion into the Habitat simulator by identifying the physical parameters of the robot. A second order dynamical model is identified from real collected trajectories. Remarkably, the policy trained in this  simulator offers better performance on real world experiments and leads to faster and smoother trajectories.

\begin{figure*}[t] \centering
\begin{tikzpicture}
        \draw (-1.5, 0) node[inner sep=0] {\includegraphics[width=0.75\linewidth]{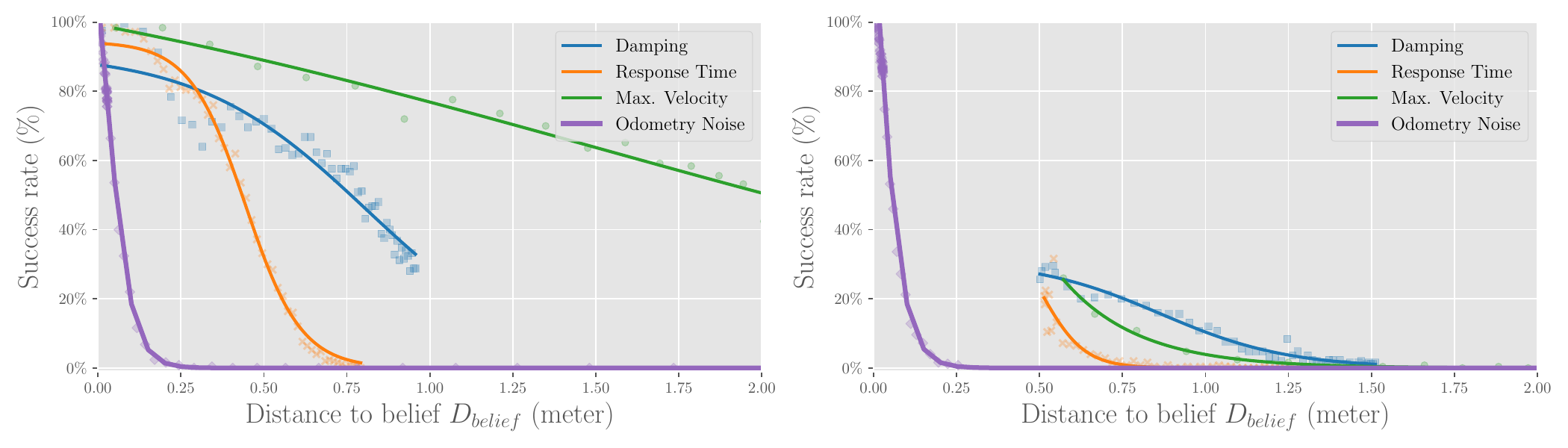}};
        \draw (7.4, 0.2) node[inner sep=0] {\includegraphics[width=0.24\linewidth]{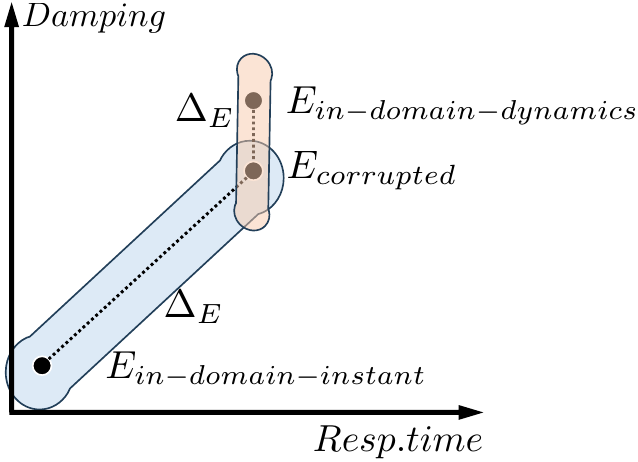}};
        \draw (-5.3, 1.7) node[inner sep=0] {\scriptsize \oursbox{D28-dynamics/Table \ref{tab:variants}(c)}};
        \draw (1.2, 1.7) node[inner sep=0] {\scriptsize \instantbox{D28-instant/Table \ref{tab:variants}(b)}};        
\end{tikzpicture}    
    \vspace*{-6mm}
    \caption{\label{fig:kalman}\simbox{\textbf{Input vs. model sensitivity}} of two different trained agents under disturbance scenarios on \epbox{HM3D/250}: \oursbox{Left: agent ``\textit{D28-dynamics}'' / Table \ref{tab:variants}(c)}, trained with the dynamical model, shows good robustness to changes in the dynamical model, but high sensibility to the odometry. \instantbox{Right: agent ``\textit{D28-instant}'' / Table \ref{tab:variants}(b)}, trained w/o dynamical model seems to overfit to the simulated ``teleportation'' behavior. The corrupted environments are the same for both and chosen as disturbances wrt. to the real dynamics, but $\Delta E$ and $D_{\text{belief}}$ are calculated to the agents' respective training environments $\rightarrow$ they are bigger for D28-instant (cf. right figure). An interactive tool with a dynamical model playground is available at
    \epbox{\href{\urlwebsitehttp}{this url}}.}
\end{figure*}

\myparagraph{Evaluation} the experimental results in the paper are color-coded to indicate the different evaluation settings:
\realbox{\textbf{(i) ``Real''}} experiments evaluate the agent trained in simulation on a real Rookie robot. This form of evaluation is the most challenging as it deals with the sim2real gap, but limits the number of episodes. We perform 20 episodes per experiment in our test building, indicated as \office, and do certain analyses on a subset of 14 episodes, indicated as \officesub. 
\simbox{\textbf{(ii) ``Simulation (+dyn. model)''}} evaluates in simulation using the Habitat simulator augmented with the realistic motion model identified from the real robot. This allows to evaluate the model on a large number of scenes and episodes while still taking into account realistic non-linear motion. We evaluate on the 2,500 episodes of the HM3D \cite{ramakrishnan2021hm3d} validation set, \epbox{HM3D/2.5k} and certain experiments are done on the 250 episodes of HM3D val-mini, \epbox{HM3D/250}. To compare performance with the real experiments, we also add experiments on a scanned and simulated version of the real episodes, e.g. \office. 

As usual, we report \textit{Success Rate} (SR), i.e., the proportion of successful episodes (terminated at ${<}0.2$m to the goal by the agent not moving and no motion ordered in sim, and ${<}1$m in real) and SPL~\citep{DBLP:journals/corr/abs-1807-06757}, i.e., SR weighted by path length,
$
\textit{SPL}{=}\sum_{i=1}^N \mathbb{I}_{\text {success}} \frac{\ell_i^*}{N\max (\ell_i, \ell_i^*)} ,
\label{eq:spl}
$
where $\ell_i$ is the agent path length and $\ell_i^*$ the GT path length. We also provide a lower bound on \textit{Success Weighted by Completion Time} (SCT)~\cite{yokoyama2021success} which weighs success by time, explicitly taking the agent's dynamics model into consideration,
$
\textit{SCT}{=}\sum_{i=1}^N S_i \frac{t_i^*}{N\max (c_i, t_i^*)}.
\label{eq:sct}
$
Here, $c_i$ is the episode time and $t^*_i$ is the optimal time on the shortest path.\footnote{as $t_i$ is an estimate in simulation, $c_i > t_i^*$ may happen.}

\myparagraph{\revision{Our modifications}} We reproduce the work from \cite{bono2024learning}, up to variations in experimental conditions inherent with evaluations in real conditions and in our own buildings. We substantially improved their performance through modifications in the engineering details, which we did not ablate, as we do not consider them to be scientific contributions: (1) we boost training from 200M to 500M env. steps, (2) we replace raw angle inputs by sin/cos projections, (3) we perform RGB data augmentations also during \textit{test time}, which alone improved SR by $\sim$15\% in the real-world setup.    

Our results are shown in Table \ref{tab:variants}. The end-to-end agent obtains excellent navigation performance with a close to optimal success rate of \reallabel{92.5\% in real world scenarios}. In particular, the policy \oursbox{(c)} learned with the simulator augmented with the dynamical model significantly outperforms the baselines \instantbox{(b)} --- same agent trained with the classical simulator w/o dynamical model, and \dfourbox{(a)} --- an agent with relative position commands $\{${\footnotesize \texttt{FORWARD 25cm}, \texttt{TURN\_LEFT $10^{\circ}$}, \texttt{TURN\_RIGHT $10^{\circ}$}, \texttt{STOP}}$\}$, trained w/o dynamical model. This confirms the necessity of adding a realistic motion model reported in \cite{bono2024learning}. In the next sections, we will provide a detailed analysis of what the agent has actually learned.

\section{Do e2e agents learn a dynamical system?}
\label{sec:dynamics}

Generally speaking, an agent has several options to estimate its current state wrt. the goal and to take actions. Two extreme solutions would be, (1) to use sensory inputs alone, or (2) to learn a latent dynamics alone and integrate it step by step. 
In the context of the studied trained navigation policies, a common (but yet untested) belief is that these agents use a combination of both, often achieved through \textit{Prediction}-\textit{Correction} steps akin to a \textit{Kalman filter} \cite{kalman_new_1960}: during the \textit{Prediction} step, the agent exploits a learned dynamical model to estimate its next state (open-loop forecasting), and during \textit{Correction} step it corrects its estimation using sensory inputs, (closed-loop improvement of the state estimate). Optimally, as the policy improves, the agent learns to better estimate its state by balancing the uncertainty of its internal dynamical model and the uncertainty in its learned sensor model. This would be, again, a generalization of the (Extended) Kalman filter, where these respective models are designed by experts and the uncertainties encoded explicitly in probability distributions. In this section, we will test this hypothesis with a series of targeted experiments, partially in simulation, partially in the real environment.

\myparagraph{\simbox{Input vs. model sensitivity}}
To quantify to which extent the trained policy is sensitive towards changes in input vs. changes in the process dynamics, we propose a new analysis method, which evaluates the agent in environments subject to normalized and thus comparable corruptions and measures the drop in success rate. We disturb the dynamical model used in the simulator by modifying the \textit{damping ratio} 
(lower values makes the dynamics closer to instantaneous at cost of larger overshoot)
the \textit{response time} (high value requires more time to reach a target velocity)
or the \textit{maximal velocity}. On the input side we corrupt by gradually increasing the gaussian noise parameters applied on the odometry sensors (\textit{mean} and \textit{std.dev}). 

\begin{wrapfigure}{r}{3.8cm}
    \vspace*{-2mm}
    \hspace*{-3mm}
    \includegraphics[width=4cm]{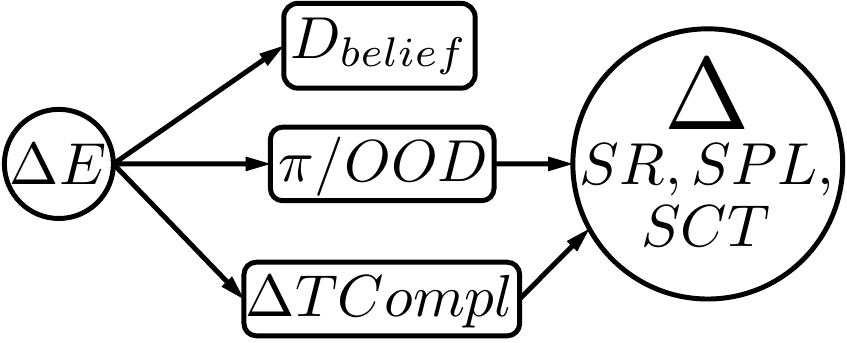}
    \vspace*{-7mm}
    \caption{\label{fig:causalgraph}The causal relationship betw. environment shifts and agent $\pi$ performance.}
    \vspace*{-4mm}
\end{wrapfigure}
Comparing changes $\Delta E$ in environment parameters is difficult, since these parameters do not have comparable ranges and have completely different meanings. In order to superpose the impacts on agents performance of these parameters in the same plots in Figure \ref{fig:kalman}, we introduce an intermediate property we call ``\textit{distance to belief}'', which quantifies the spatial distance (expressed in \textit{meters}) of the agent in the corrupted environment compared to its behavior in the training environment. To this end, we collected a set of 1,000 5s-long sequences of actions and measured average spatial distance of the agent path running in both versions, before and after $\Delta E$ (see fig. \ref{fig:kalman_metric} and more details in appendix \ref{ap:calculation_d_belief}).

Figure \ref{fig:causalgraph} shows the underlying causal graph: a change $\Delta E$ translates into a measure $D_{belief}$, comparable between parameters. It also gives rise to a change in navigation performance, through two mediators: the out-of-distribution situation of the agent, ``$\pi/OOD$'', and the eventual difference in task complexity, $\Delta{T}Compl$. 

We report our results in Figure \ref{fig:kalman}: for both agents, trained w/ (left) or w/o (right) dynamical model, the largest sensitivity can be observed for odometry inputs, indicating that the agent significantly relies on its sensing capabilities to correct a state potentially estimated by open-loop forecasting. For the agent trained w/ dynamical model, changes in the parameters of the dynamics have a large effect, in particular response time and damping, almost comparable to changes in input odometry. Performance is low for the agent trained w/o dynamics, it seems to overfit to the ``teleportation'' seen in simulation. The gap between $0{<}D_{belief}{<}0.5$ is related to the OOD situation, i.e. the minimal difference between in-domain dynamics (here, instant and constant velocities) and corrupted dynamics (more details in Sec. \ref{app:kalman_details}).

\begin{figure}[t] \centering
    \includegraphics[width=\linewidth]{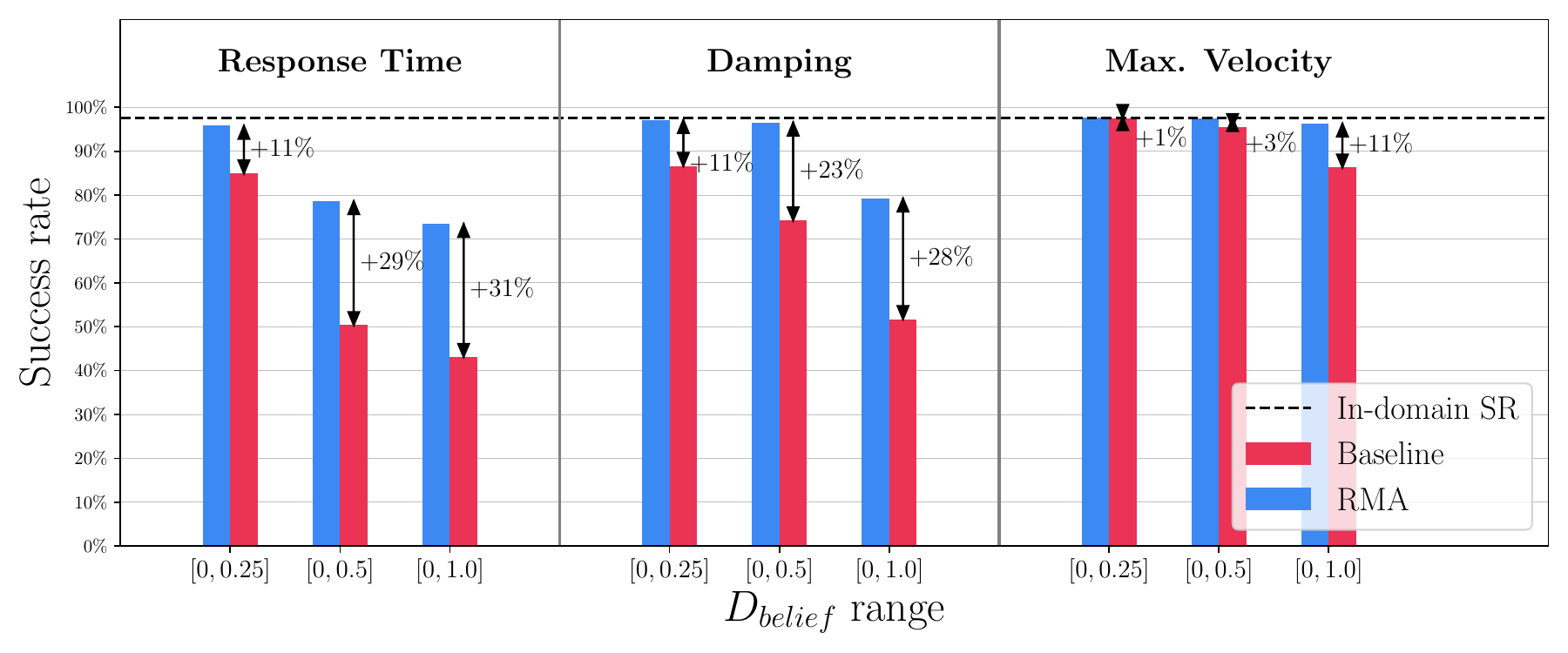}
    \vspace*{-8mm}
    \caption{\label{fig:rma}\textbf{Learning robustness:} inspired by RMA \cite{RMA2021}, we train a policy capable of adapting to changes $\Delta E$ in env parameters .}  
    \vspace*{-5mm}
\end{figure}

\myparagraph{\simbox{Overcoming the impact of $\Delta E$}} Inspired by RMA \cite{RMA2021}, we train a new policy with a simulator which randomly samples environment changes $\Delta E$ during each training episode, aiming to train a policy capable of adapting to them. We encode the environment parameters into an embedding vector $\mathbf{e}_E$, which is passed to the policy as input, leading to a new architecture $\mathbf{a}_t{=}\pi(\mathbf{h}_t, \mathbf{e}_E)$, replacing eq. (\ref{eq:policy}). Figure \ref{fig:rma} shows the drop in performance between in-domain and OOD (applying $\Delta E$) and then the gain back in performance by the RMA-augmented policy, which has access to the embedded ground-truth environment parameters. This provides evidence, that changes in dynamics have an important impact on agent performance, and that the agent can adapt if it has access to them or can estimate them.

\begin{figure}[t]
    \centering
    \begin{overpic}[width=0.4\textwidth]{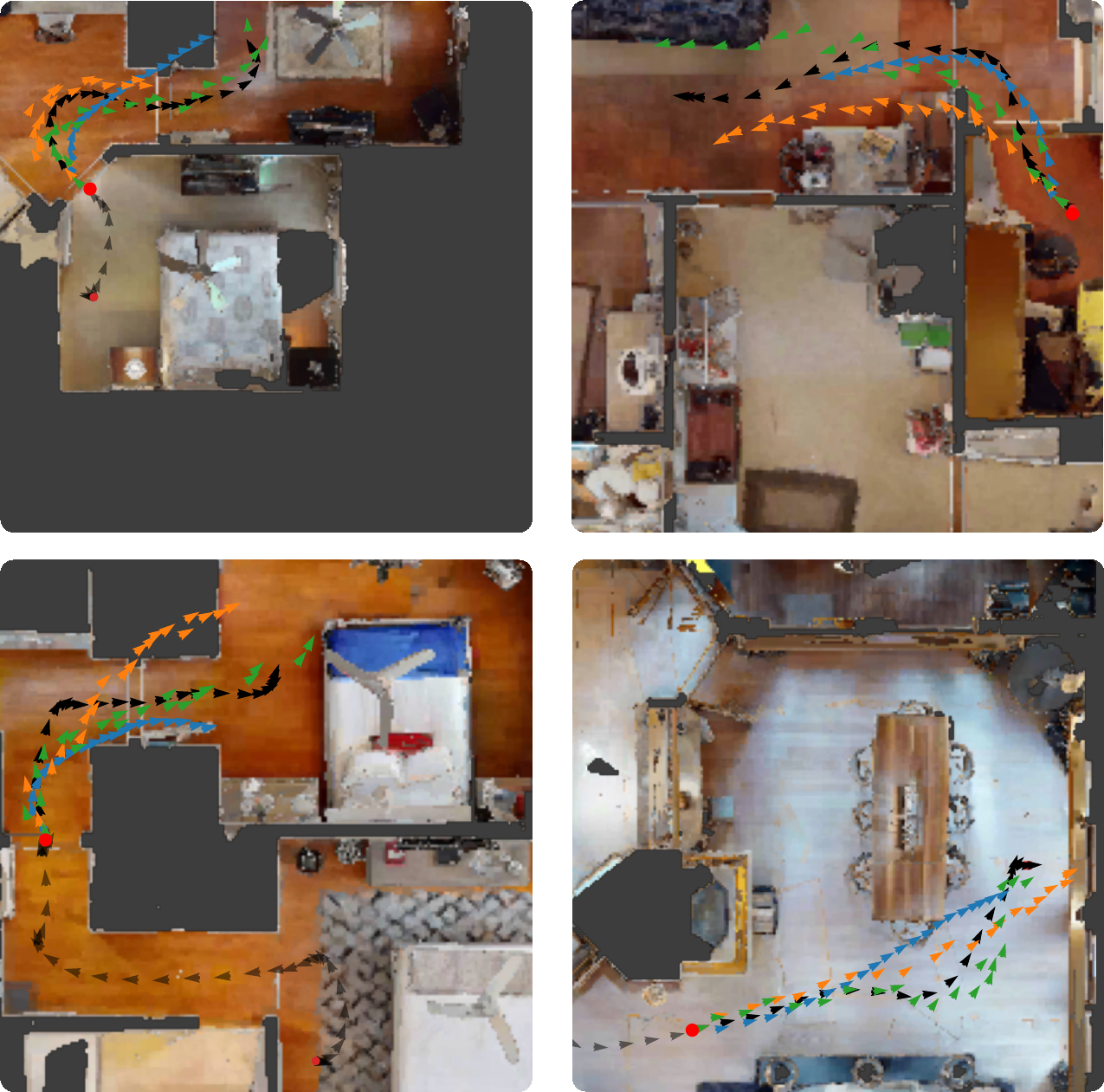}
        \put(35,45){
            \adjustbox{scale=0.8, center=2.5cm,raise=0.5cm}{
                \footnotesize
                \begin{tabular}{|c|c|c|}
                    \specialrule{1pt}{0pt}{0pt}
                    \rowcolor{white}Probing Model & Pos. error & Ang. error \\
                    \specialrule{1pt}{0pt}{0pt}
                    \rowcolor[HTML]{6cacd9}Agent-GRU   & 0.564   & 0.342   \\
                    \rowcolor[HTML]{fab06e}Linear   & 0.767   & 0,3856   \\
                    \rowcolor[HTML]{67bf67}Linear + prev-action &  \textbf{0.441}   & \textbf{0.205}   \\ \specialrule{1pt}{0pt}{0pt}
                \end{tabular}
            }
        }
    \end{overpic}
    \vspace*{-2mm}
    \caption{\label{fig:probingdynamics}\simbox{\textbf{Probing dynamics:}} using the agent's own internal dynamics, it is capable of predicting future pose from time $t$ (red dot \textcolor{red}{\ding{108}}) through a trained probing network. Black/gray: GT trajectory.}    
    \vspace*{-5mm}
\end{figure}

\begin{figure*}
    \centering
    \begin{subfigure}{0.5\textwidth}
        \includegraphics[width=\textwidth]{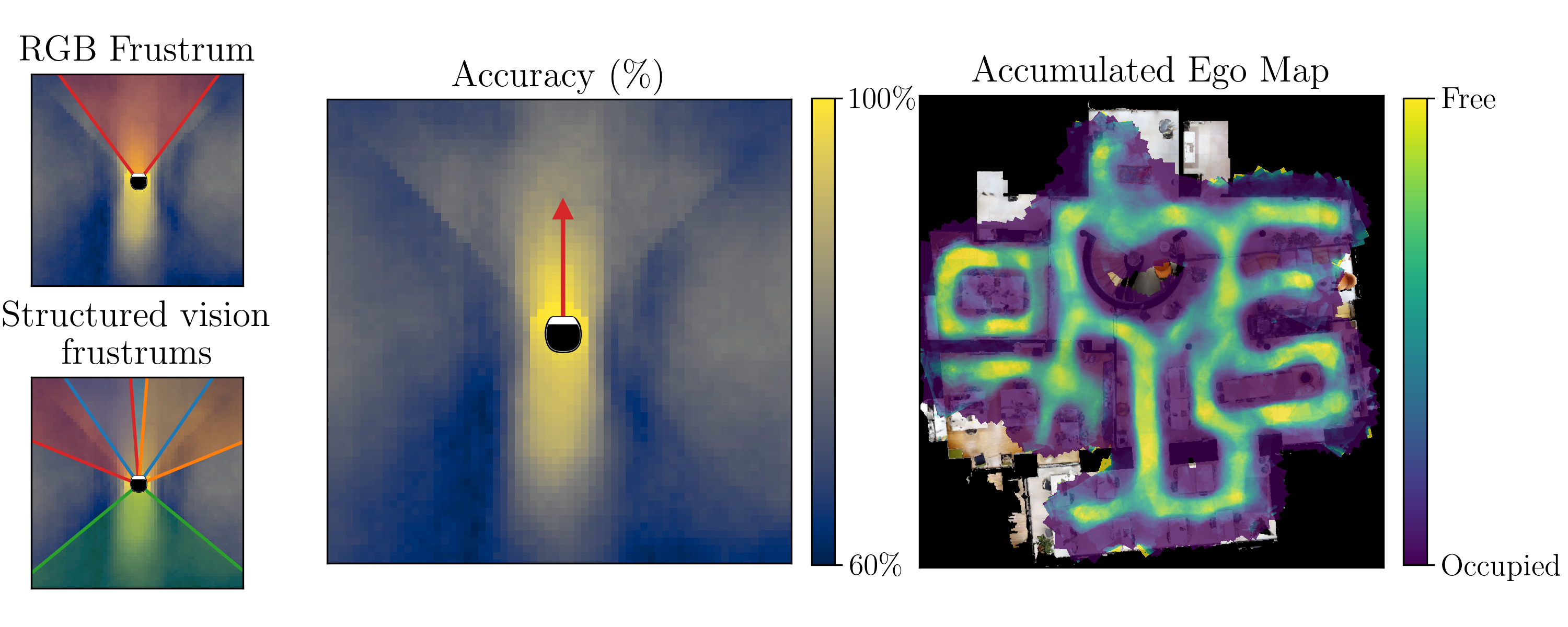}
        \caption{\simbox{Maps probed from simulated states $\mathbf{h}_t$} on \office}
    \end{subfigure}
    \begin{subfigure}{0.4\textwidth}
        \includegraphics[width=\textwidth]{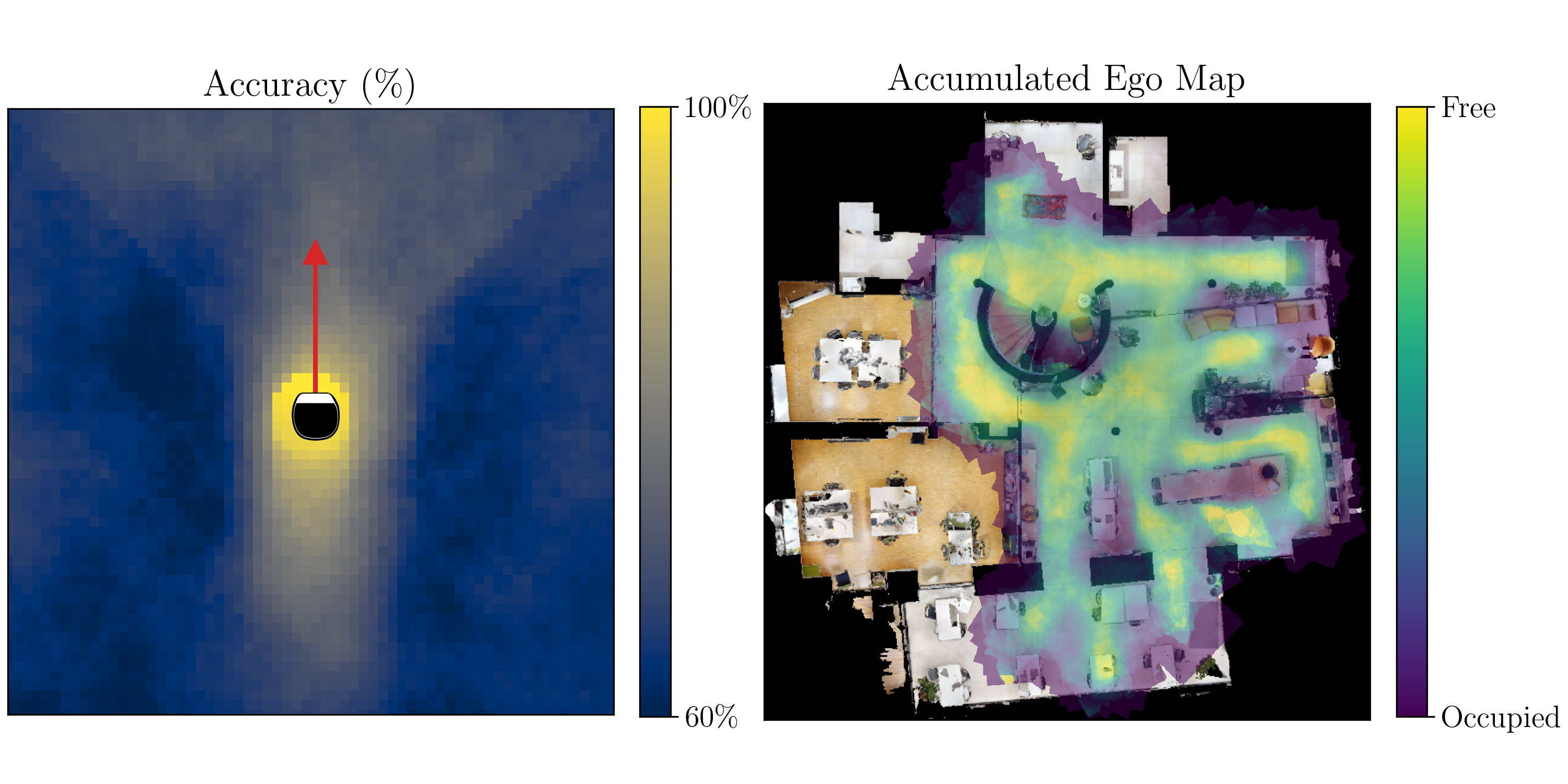}
        \caption{\realbox{Maps probed from real states $\mathbf{h}_t$} on \officesub \\ \tiny{*Accuracy computed by ignoring furniture changes and localization error}}
    \end{subfigure}
    \vspace*{-3mm}    
    \caption{\label{fig:occupancyprobing}\textbf{Occupancy probing} -- we trained a probing network to reconstruct the local occupancy map (in a 3m square around the robot) on HM3D. For simulated (a) and real (b) trajectories, we compute the accuracy (left) of the predicted occupancy map exhibiting large error in the dead zone of the structured vision sensors (left most figures). We also accumulate the probed maps over multiple episodes and super-imposed the result with the scene actual map, using onboard pose estimates for real data (right).} 
    \vspace*{-4mm}
\end{figure*}

\myparagraph{\simbox{Probing dynamics}} We further analyze the agent by directly probing whether it is capable to predict its future pose. We generate a dataset of pairs of hidden states $\mathbf{h}_t$ and agent pose $\mathbf{p}_t$ from trajectories on the HM3D train split and train a model to predict future pose $\mathbf{p}_{t+\tau}$ from $\mathbf{h}_t$. We train two variants, each of which is a separate network for each horizon $i$ predicting $\mathbf{p}_{t+i}$. One is a linear model, the second one has access to the goal and prev. action $\mathbf{a}_{t+i-1}$,
\begin{equation*}
    \begin{array}{llr}
    \mathbf{p}_{t+i} &= \text{Linear}_i\big(\mathbf{h}_{t}\big) &\rem{Linear} \\ 
    \mathbf{p}_{t+i} & = \text{Linear}_i\big(\mathbf{h}_{t}, \text{MLP}(\mathbf{a}_{t+i-1}, \mathbf{g})\big) &\rem{Linear+p-act}
    \end{array}
\end{equation*}
We'd like to insist, that \textit{neither of the models have access to observations beyond time step $t$}. Visualizations on 4 trajectories and numerical errors from scenes unseen during training are given in Figure \ref{fig:probingdynamics}. The \textit{linear} variant (\textcolor{probinglinear}{orange curve}) shows quite low prediction error and good quality behavior in a reasonable horizon but quickly seems to break down for larger horizons. The \textit{linear + prev-action} variant (\textcolor{probinglinearpact}{green curve}) has lower error and shows better long-term behavior, indicating that providing some elementary information on the plan (the goal and a single previous action) to the probing network is helpful. We also explore a version, which uses the latent dynamics of the agent itself rolled out auto-regressively into the future (\textcolor{probinggru}{blue curve}). It tests whether the recurrent update the agent did over its past, as given by eq. (\ref{eq:gru}), can also be used to rollout the latent state auto-regressively into the future for prediction, but without future observations (see \suppm{sec:probingdynamics} for more details and equations). Needless to say that the policy does not have access to such a model and as expected, the results are lukewarm. 
The numerical results in the inlet of Figure \ref{fig:probingdynamics} are prediction errors for 39k trajectories in unseen scenes given for 20 steps, ie. $\sim$6s into the future. A mean of error 0.76m after 6s is arguably quite low, given that in many cases an accurate prediction cannot be done without future observations, as can be easily seen in the upper left trajectory of Figure \ref{fig:occupancyprobing}. We argue for the presence of short-to-medium horizon dynamics in the latent representation $\mathbf{h}_t$.

\myparagraph{\realbox{Max velocity}} The agent has been trained for a maximum velocity of 1m/s, and in Table \ref{tab:real_max_vel} we test the impact of clipping its maximum speed on the real robot. Limiting speed to 0.7m/s is beneficial: we conjecture that the disadvantage of being OOD wrt. to the training speed is compensated by the slightly easier task. Decreasing the speed limit further is not helpful anymore. We set the limitation to 0.7m/s as the default behavior in all our experiments. Retraining the agent with this limitation could potentially increase performance, and will be left for future work. \textit{In all tables, lines marked by ``(def)'' indicate the default parameter settings used in the rest of the experiments}.

\myparagraph{\realbox{Observation to decision delay}} The decision loop of 3Hz provides 333ms to perform observation collection, pre-processing and network forward pass, which our onboard \textit{Nvidia Jetson AGX Orin} executes in \env{}100ms. We exploit the unused \env{}230ms in real world experiments, by taking decisions \textit{faster} than in the 333ms done during training, cf. Table \ref{tab:real_delay}. The combined effects of being OOD and deciding faster on less ``stale'' observations does not seem to have significant effects on navigation performance.

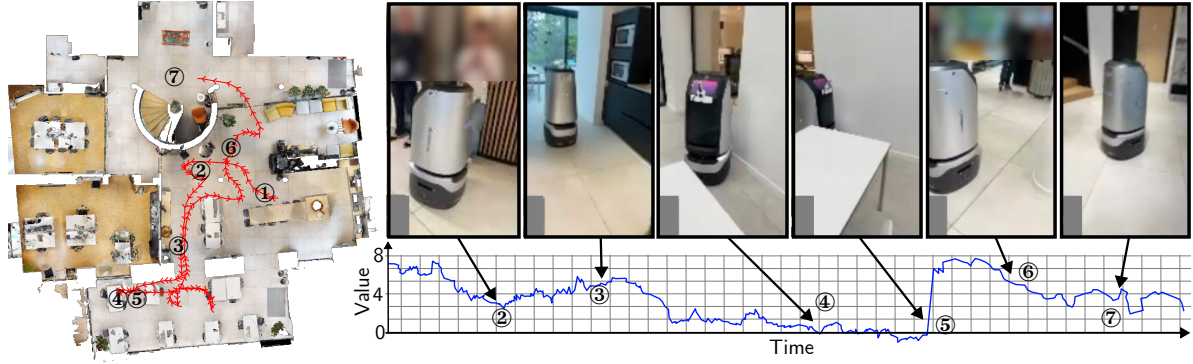
\begin{figure*} \centering
\begin{tikzpicture}
        \draw (0, 0) node[inner sep=0] {\includegraphics[width=0.9\linewidth]{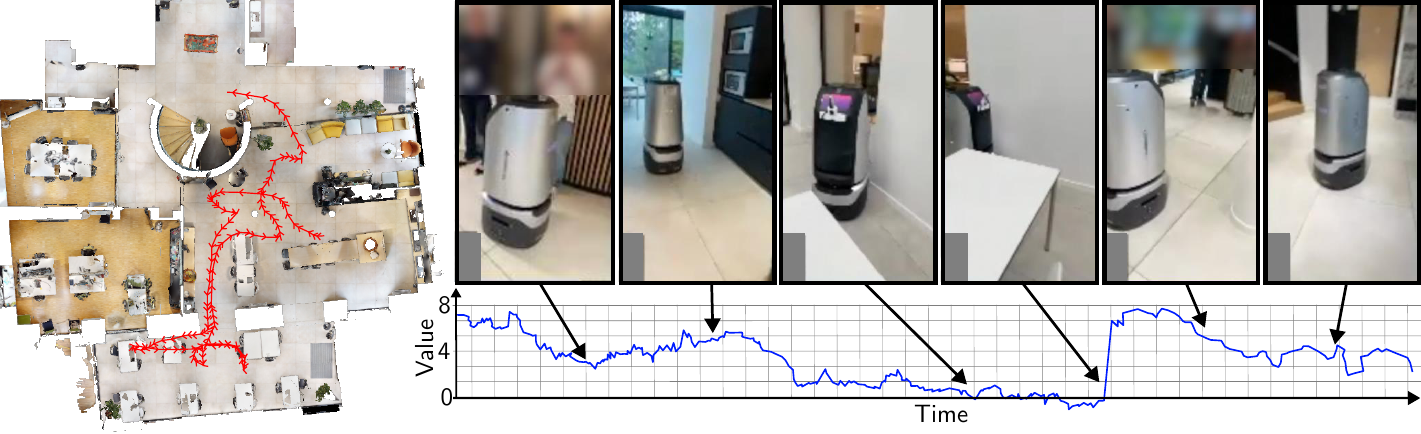}};
        \draw (-1.3, -1.8) node {\ding{173}};
        \draw (0, -1.5) node {\ding{174}};
        \draw (3.0, -1.6) node {\ding{175}};
        \draw (4.6, -1.9) node {\ding{176}};
        \draw (5.7, -1.2) node {\ding{177}};
        \draw (6.8, -1.8) node {\ding{178}};
        \draw (-4.45, -0.15) node {\ding{172}};
        \draw (-5.3, 0.15) node {\ding{173}};
        \draw (-5.6, -0.85) node {\ding{174}};
        \draw (-6.4, -1.55) node {\ding{175}};
        \draw (-6.15, -1.55) node {\ding{176}};
        \draw (-4.9, 0.45) node {\ding{177}};
        \draw (-5.65, 1.4) node {\ding{178}};
        \end{tikzpicture}
\vspace*{-3mm}
\caption{\label{fig:irene}\realbox{\textbf{Long horizon estimate: visualization of the value estimates by the PPO critic}} --- in this episode the agent started at pos \ding{172} with the goal at pos \ding{178}. Two paths are possible, the agent chooses the slightly longer one passing through the door left of pos \ding{173}, which is blocked by people. It decides to take an alternative route south towards \ding{174}, the value estimate dropping. At \ding{175} the agent tries circumvent the situation by going north, but can't find a path through the glass panels, the value estimate is now negative. At \ding{176}, finally, the agent decides to abandon this strategy and seems to re-plan. The value estimate immediately pikes, the agent seems to anticipate a long series of positive reward. At \ding{177} we are back to the blocked position with similar reward, and the agent now decides to try the other door. At \ding{178} we reach the goal, and the value estimate converges to 2.5, equal to the final reward for a successful episode \realbox{(full video in the suppl. material).}}
\vspace*{-3mm}
\end{figure*}

\begin{table*}[t]
\small
\begin{minipage}[t]{0.32\linewidth}
    \begin{tabularx}{\linewidth}{L|RRR}
       \specialrule{1pt}{0pt}{0pt}          
         \textbf{Delay (ms)} & 
         \cellcolor{green!50} \textbf{SR} & 
         \cellcolor{green!50} \textbf{SPL} &
         \cellcolor{green!50} \textbf{SCT} \\
         \specialrule{1pt}{0pt}{0pt}
         325 & 100.0 & 66.5 & 24.1 \\
         270 & 100.0 & 61.1 & 23.8 \\
         225 & ~~92.9 & 57.7 & 22.0 \\
         190 & ~~92.9 & 62.5 & 24.6 \\
         150 (def) & 100.0 & 63.3 & 25.9 \\
         130 & 100.0 & 66.0 & 24.5 \\
         \specialrule{1pt}{0pt}{0pt}
    \end{tabularx}
    \caption{\label{tab:real_delay}\realbox{\textbf{Obs. to decision delay}} during testing on real, \officesub{} --- compared to 333ms seen during training (3 Hz): performance is not sensitive to OOD conditions.}
\end{minipage}
\hfill
\begin{minipage}[t]{0.32\linewidth}
    \begin{tabularx}{\linewidth}{L|RRR}
       \specialrule{1pt}{0pt}{0pt}          
         \textbf{Max. Vel.} & 
         \cellcolor{green!50} \textbf{SR} & 
         \cellcolor{green!50} \textbf{SPL} &
         \cellcolor{green!50} \textbf{SCT} \\
         \specialrule{1pt}{0pt}{0pt}         
         100\% & 85.0 & 54.0 &20.9\\
         ~~70\% (def) & 100.0 & 63.3 & 25.9 \\
         ~~50\% & 85.0 & 66.1 & 23.3\\         
         \specialrule{1pt}{0pt}{0pt}
    \end{tabularx}
    \caption{\label{tab:real_max_vel}\realbox{\textbf{Impact of maximum velocity}}: in simulation, the agent is trained with a maximum speed of \SI{1}{\meter/\second}. We evaluate the policy in real world with different maximum speeds, on \officesub.}
\end{minipage}
\hfill
\begin{minipage}[t]{0.32\linewidth}
    \begin{tabularx}{\linewidth}{LL|RRR}
       \specialrule{1pt}{0pt}{0pt}          
         \textbf{Periodic} &  \textbf{$<$2m} &
         \cellcolor{green!50} \textbf{SR} & 
         \cellcolor{green!50} \textbf{SPL} &
         \cellcolor{green!50} \textbf{SCT} \\
         \specialrule{1pt}{0pt}{0pt}
         ~~~~~\no  & \no  & 40.0 & 32.3 & 10.7\\
         ~~~~~\no (def) & \yes (def) & 100.0 & 63.3 & 25.9\\
         30 sec &\yes& 85.0 & 58.9 & 23.7 \\
         10 sec &\yes& 80.0 & 59.9 & 24.5 \\         
         ~~3 sec &\yes& 75.0 & 55.9 & 18.8 \\         
         \specialrule{1pt}{0pt}{0pt}
    \end{tabularx}
    \caption{\label{tab:resend}\realbox{\textbf{Ablating agent memory}}: periodically zeroing $\mathbf{h}_t$, or at the end of the episode when dist. to goal $<$2m \officesub.}
\end{minipage}
\vspace{-5mm}
\end{table*}

\section{Does e2e training lead to planning?}
\label{sec:planning}

The agent analyzed in our work does not have an explicit baked-in inductive bias for planning, and we raise the question whether planning emerges through RL training alone. 

\myparagraph{\simbox{Probing plans}}
In section \ref{sec:dynamics} we introduced probing future pose $\mathbf{p}_{t+\tau}$ from $\mathbf{h}_t$. The low prediction errors given in Figure \ref{fig:probingdynamics} on trajectories of the unseen HM3D validation scenes can only be achieved if the agent
(i) is able to leverage a learned notion of dynamics, as argued, and (ii) has a latent plan in its hidden state, as predicting the future pose also requires access to future actions. The visualizations also provide evidence of short-to-medium horizon plans, eg. direction changes  aligned between GT and prediction.

\myparagraph{\realbox{Analysis of the value estimate}} As the agent had been trained with PPO, its learned critic can provide an estimate of the value function. While the value estimate does not provide a long-horizon plan of the agent, it does provide the agent's estimate of the expected cumulated future rewards, and as such provides \textit{some} indication of what the agent plans to achieve. In Figure \ref{fig:irene}, we provide a post-hoc analysis of a single episode collected during a demonstration of the agent in a crowded scenario. The episode provides evidence, that navigation strategies in the form of choices of paths are taken, tested, and rejected to be replaced by better options. In particular, these choices seem to have an effect on the value estimate. Abandoning a navigation option for a more promising one increases the value estimate, as the agent now expects a higher future return. This gives some evidence that the agent has an idea of where it stands in a plan structured on the level of paths and that its estimate of success goes beyond the effect of the next action.

\myparagraph{\realbox{Comparison with an expert planner}} We introduce an expert in the form of the \textit{Fast Marching Square} method~\cite{gomez2013FM2} working on a floor-plan of our test building, and which produces a cost function $\mathcal{C}(\mathbf{p}_t, \mathbf{a}_t)$ taking into consideration the effect of the agent's action  $\mathbf{a}_t$ on the geodesic distance from position $\mathbf{p}_t$ to the target. It also considers the alignment of the agent to the gradient of the geodesic distance, as well as velocity term that encourages high velocity when far from the target and a low velocity when approaching it in order to stop correctly ---  cf. \suppm{sec:heatmap} for details.

\begin{figure}
    \begin{tikzpicture}
        \draw (0, 0) node[anchor=west, inner sep=0] {\includegraphics[width=\linewidth]{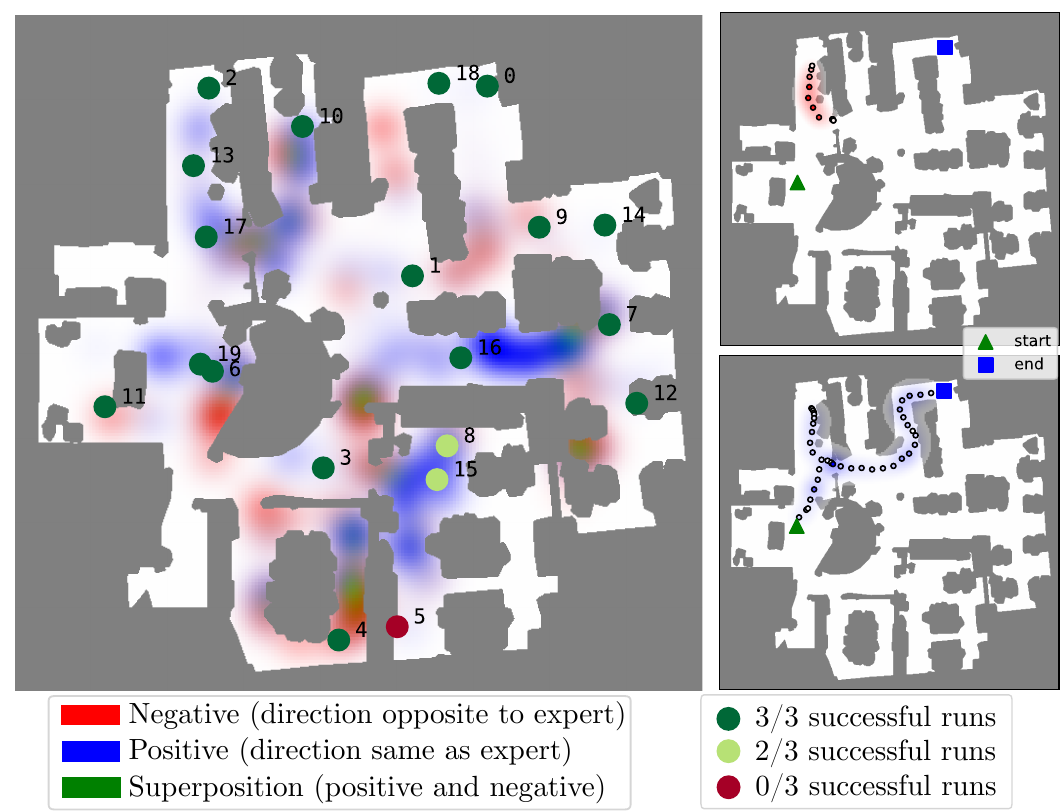}};        
        \draw (1.5, 2.9) node[anchor=west, inner sep=0] {\scriptsize\oursbox{3$\times$20 episodes, pos+neg}};  
        \draw (6.2, 0.8) node[anchor=west, inner sep=0] {\scriptsize\oursbox{1 episode, neg}};
        \draw (6.2, -2) node[anchor=west, inner sep=0] {\scriptsize\oursbox{1 episode, pos}};
    \end{tikzpicture}
    \vspace*{-5mm}
    \caption{\label{fig:heatmaps}\realbox{\textbf{Planning quality map}} --- we visualize the difference in cost, as expressed by eq. (\ref{eq:cost}) caused by the actions of the agent, on 60 episodes (Top), and a single episode (Bottom) for negative (left) and postive (right) actions.}
    \vspace*{-5mm}
\end{figure}

We compare trajectories of the agent in the real environment with expert trajectories using the cost function to estimate a \textit{planning quality measure} for each time step $t$ as,
\begin{equation}
    M(t) =\mathcal{C}(\mathbf{p}_{t+1}, \mathbf{a}_{t+1}) - \mathcal{C}(\mathbf{p}_t, \textbf{a}_t).
    \label{eq:cost}
\end{equation}
We translate this into a heatmap using kernel density estimation. To be more precise, we separately estimate positive and negative values and superpose them, as seen in Figure \ref{fig:heatmaps} (top) for the 60 episodes of 3 experiments from Table \ref{tab:variants}(c). Figure \ref{fig:heatmaps} (bottom) shows the positive and negative actions of a single episode. Episodes are defined as navigation between consecutive goals, eg. 0$\rightarrow$1.

The heatmap indicates compatibility with expert decisions near bottleneck areas like doors (near goal numbers 16, 8, 15). The episode from 4$\rightarrow$5 is particularly difficult due to its high difference between Euclidean distance and geodesic path: it always failed due to problems in long-term planning, corroborated by the bad decisions visible in the heatmap. Other complicated areas are narrow passages (eg. near goal 1), where the agent fails to execute its plans.

\section{Do agents use episodic memory?}
\label{sec:memory}

The agent's latent representation $\mathbf{h}_t$ is a compression of its history of observations trained to keep the most relevant information for the task. We probe its impact on navigation.

\myparagraph{\realbox{Ablating memory}} In Table \ref{tab:resend} we test the agent in the real env. by zeroing $\mathbf{h}_t$ periodically every 30s, 10s or 3s. Note, that this requires resetting episode start such that the goal is redefined, since the agent potentially uses its latent state to translate the static goal to an ego-centric frame (see \suppm{sec:zeroing}). We see a steady decrease of agent performance when the frequency of memory ablations is increased, with a drop of 25\% in SR when memory is zeroed every 3s. As expected, the agent loses its capacity to remember which parts of the scene have been explored and gets stuck. 

In our experiments we zero memory when the onboard sensors estimate that the agent is closer than 2m from the goal. Without, performance drops to 40\% SR (first line in Table \ref{tab:resend}). We link this behavior to the task setting as static point goal, where the goal coordinates fed to the agent are constant and wrt. to the episode start. Refreshing memory helps to better deal with the very last meter to the goal and prevents its from prematurely deciding to stop.

\myparagraph{\realbox{Memory and scene structure}} We trained a probing network to predict a local occupancy map of $3{\times}3m$ centered on the agent's position from the agent memory $\mathbf{h}_t$. We trained on the HM3D train split and show results on our test building in simulation and on real data in Figure \ref{fig:occupancyprobing}. We also integrated all these probing results from the 14 episodes of \officesub{} and super-imposed them over the map of the building, both in simulation and using probes from the \reallabel{real data (inputs and memory)}. For the latter, we used the onboard pose estimates from the agent. The occupancy predictions align very well with the real occupancy of the scene, also in difficult areas with doors and transparent walls.

\section{Sensitivity of observation types}
\label{sec:inputs}

\begin{figure}[t] \centering    
\begin{tikzpicture}
        \draw (-5, 0) node[inner sep=0] {\includegraphics[width=0.49\linewidth]{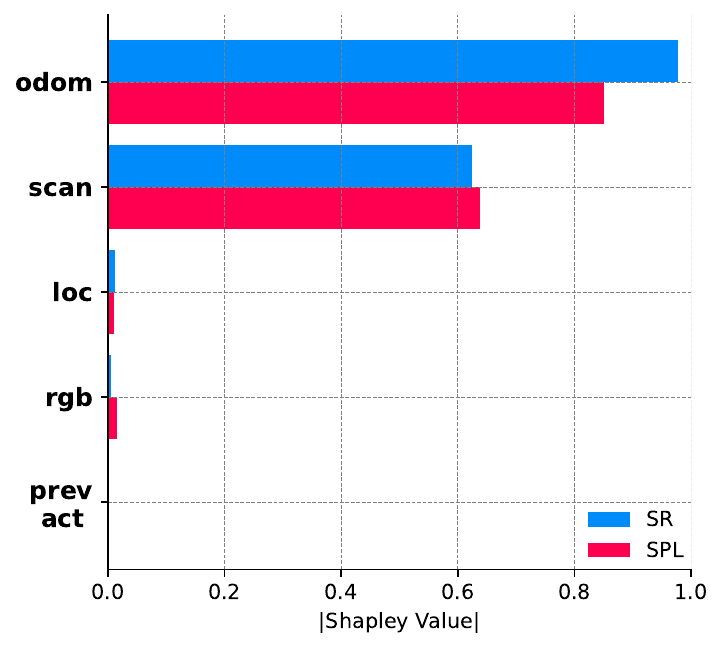}};
        \draw (-1, 0) node[inner sep=0] {\includegraphics[width=0.49\linewidth]{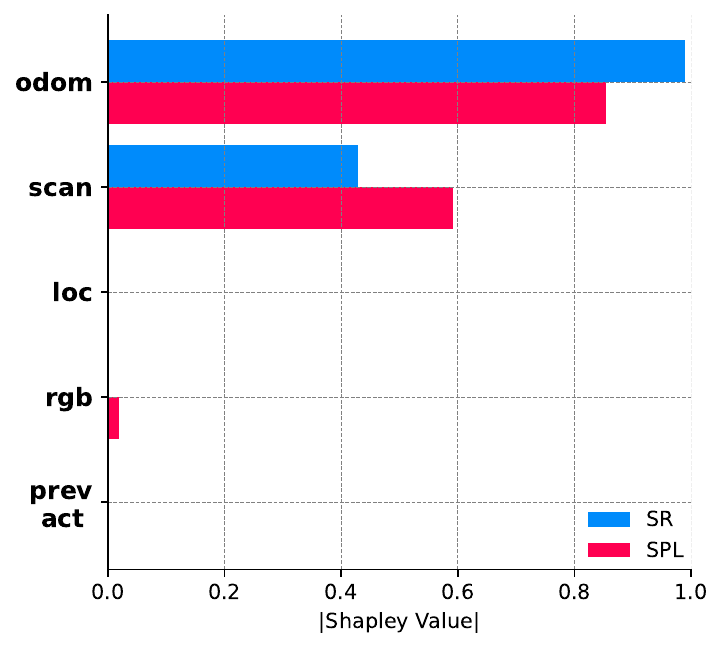}};
        \draw (-4.5, 1.9) node[inner sep=0] {\scriptsize \oursbox{D28-dynamics/Table \ref{tab:variants}(c)}};
        \draw (-1, 1.9) node[inner sep=0] {\scriptsize \instantbox{D28-instant/Table \ref{tab:variants}(b)}};
\end{tikzpicture}
    \vspace*{-3mm}
    \caption{\label{fig:shapeley}\textbf{Importance of different observations} with Shapely features, for \oursbox{D28-dynamics} (left) and \instantbox{D28-instant} (right).}
    \vspace*{-5mm}
\end{figure}

\myparagraph{\simbox{Post-hoc analysis}} We performed Shapley value analysis \cite{shapley1951} to assess the impact of the various input modalities on navigation performance. Based on cooperative game theory, it helps to fairly allocate a payoff among different players of a game. In our case, the players are different inputs and the Shapley value quantifies the contribution of each input to the agent's SR and SPL. 
To this end, we created a `background' dataset with alternative inputs derived from previously unseen scenes and systematically perturbed one input modality at every step.
This was conducted separately for odometry, localization, RGB, scan, and previous action inputs. Fig. \ref{fig:shapeley} shows, that the agent heavily relies on odometry and scans, while RGB, localization, and previous actions contribute significantly less to the final outcome.

\myparagraph{\realbox{Visual localization}} We explore replacing the pose input $\hat{\mathbf{p}}^a_t$ calculated with AMCL from the Lidar input with visual localization calculated by the \reallabel{real robot's onboard sensors}. We use visual localization with repeatable keypoints R2D2 \cite{R2D2_2019} on images retrieved from a dataset of observations \cite{lee2021largescalelocalizationdatasetscrowded} --- cf. \suppm{sec:vloc}. Results in Table \ref{tab:vloc} show that visual localization underperforms as measured by navigation metrics, which we link to failures to call STOP precisely enough. As in (static) PointNav the goal is not specified visually but with episode centric coordinates, localization is used heavily for the last meters, and localization from visual input does not emerge from the RL loss alone.

\begin{table}[t]
    \centering
    \small
    \begin{tabularx}{\linewidth}{L|RRR}
       \specialrule{1pt}{0pt}{0pt}          
         \textbf{Pose input $\hat{\mathbf{p}}^a_t$} & 
         \cellcolor{green!50} \textbf{SR(\%)} & 
         \cellcolor{green!50} \textbf{SPL(\%)} &
         \cellcolor{green!50} \textbf{SCT(\%)} \\
         \specialrule{1pt}{0pt}{0pt}
         Visual local. w. R2D2 \cite{R2D2_2019} & ~~42.9 & 23.2 & ~~7.8 \\
         AMCL from 1D Lidar (def) & 100.0 & 63.3 & 25.9 \\         
         \specialrule{1pt}{0pt}{0pt}
    \end{tabularx}
    \vspace*{-3mm}
    \caption{\label{tab:vloc}\realbox{\textbf{Visual localization}}: replacing ROS/AMCL from 1D-Lidar input with R2D2+retrieval from RGB  \cite{R2D2_2019} \officesub{}.}
    \vspace*{-5mm}
\end{table}

\section{Discussion and conclusion}

In this paper we raised the question whether and what kind of models emerge in a navigation agent trained with model-free RL, in particular when trained on realistic motion. Our findings showed, that some limited information on planning is present in the end-to-end trained memory representation, which can be exploited by a linear probe to predict poses in a short-to-medium time horizon with a reasonable precision. This capacity emerges although planning has not been baked into the agent --- the architecture is purely recurrent and updates a hidden memory over time up to the present, without performing any rollouts into the future, as for instance MPC or TD-MPC like models do \cite{hansen2024tdmpc2}. 

It can be argued, that the actor-critic RL algorithm we used for training, PPO~\cite{schulman2017proximal}, did provide some form of long-horizon training signal through the loss of its value estimate (\textit{Generalized Advantage Estimation}~\cite{schulman2018GAE}). Our instance based post-hoc analyzes gave some evidence, that the PPO value critic is indeed capable of providing value estimates, which are linked to long-term plans of the agent, as can be seen by sharp discontinuities in the value estimate during changes of strategy and moments of observed ``re-planning''. This was confirmed in real world settings, transferred from simulation. The elements of a long-term plan seem to be present in the agent's memory, as evidenced by the simple form of the PPO-value estimate --- a non-linear projection from the hidden state. Ablating the memory further confirmed its importance, leading to metrics drop as the agent can't retain information on unsuccessful trajectories. We again stress, that a large part of these experiments have been done on a real robot in \numepisodes{} episodes.

Further experiments probed the same agent memory for the presence of a precise fine-grained motion model, which was confirmed by numerical results up to a reasonable precision. This arguable comes to no surprise, as the control of a fast-moving robot with a low-frequency decision loop of 3 Hz only requires some form of anticipation of the robots future pose. 
We proposed a new sensitivity analysis that compares the effects of OOD behavior wrt. varying dynamics and odometry inputs in a consistent manner. It showed the reliance of the agent on odometry input, but also the strong positive impact of training the agent on realistic motion. This method also suggests that the agent has learned a Kalman-like procedure of prediction and correction steps.

\myparagraph{What we did not see} our experiments failed to provide evidence for correct long-term plans of the agent. We noticed a form of ``tunnel vision'', where agent made strategical choices for long-horizon paths, which a humans could have quickly evaluated as fruitless using their high-level geometric understanding and situation awareness --- see \suppm{sec:tunnelvision} for an example. We argue that additional work on large-scale training of geometric foundation models could provide further improvements in navigation and Embodied AI when used as pre-trained visual encoders.

{
    \small
    \bibliographystyle{ieeenat_fullname}
    \bibliography{root}
}

\clearpage
\setcounter{page}{1}
\appendix

\twocolumn[{%
\renewcommand\twocolumn[1][]{#1}%
\maketitlesupplementary
\includegraphics[width=\linewidth]{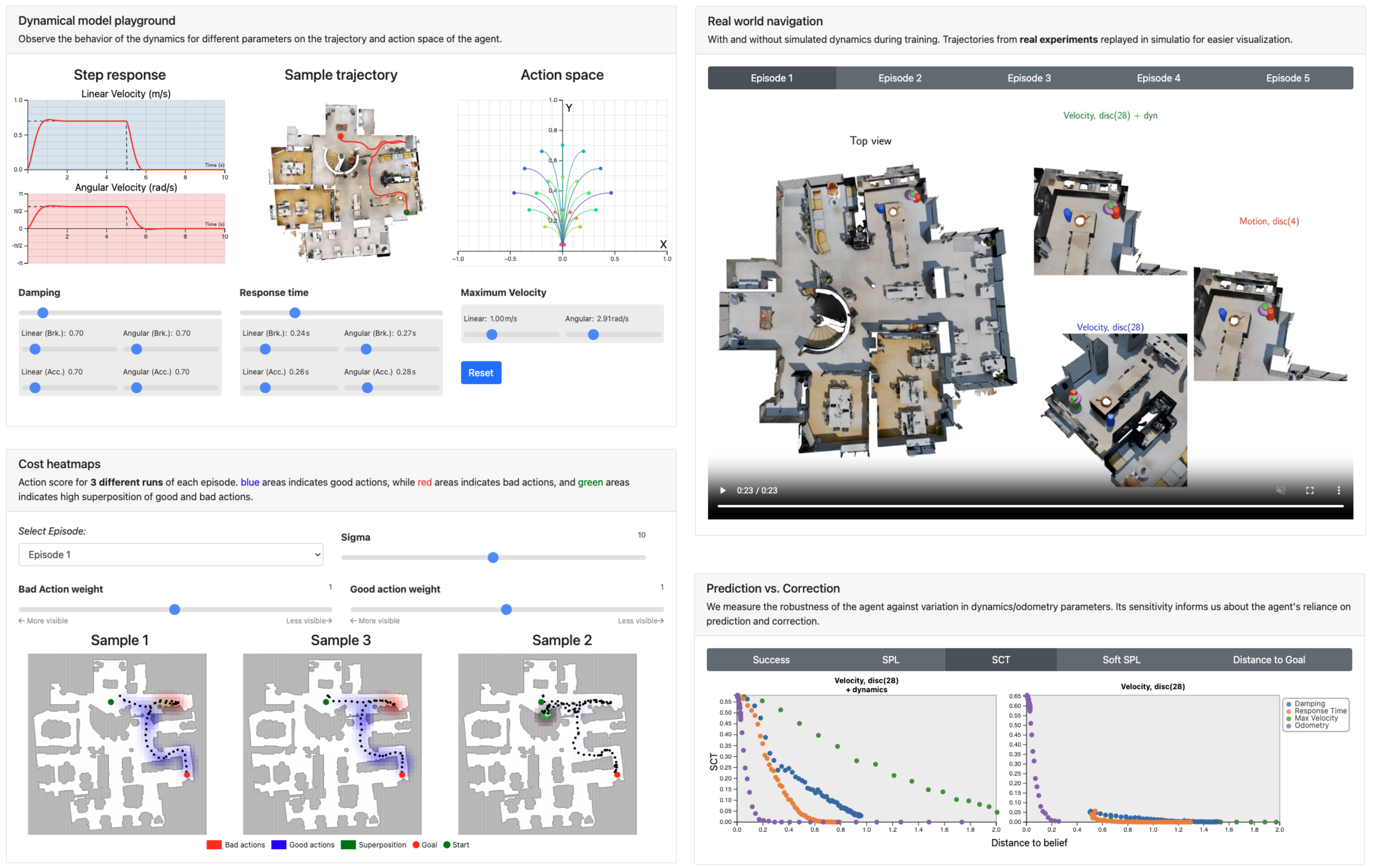}
\captionof{figure}{\label{fig:website}We created an interactive website featuring several data visualization tools to help illustrating our findings, such as a real-time dynamical model similar to the one used in the simulator, allowing to observe directly the impact of each parameter on the behavior of the robot.  \epbox{\href{\urlwebsitehttp}{Project Page}}}.   
\vspace*{4mm}
}]

\mysection{0}{Interactive website}
\label{sec:website}

We developed an interactive website to support our findings and help better visualize the results of our experiments. In particular, our project page features an interactive second order dynamical model similar to the one implemented in the simulator. Several sliders control the value of physical parameters from the model, and the animated figure displays the impact on the step response, the trajectory and the action space in real-time. We also replayed real episodes from the different methods in Table \ref{tab:variants} synchronized on the same scene to better compare them --- although these episodes are replayed in the simulator, these were recorded with the agent \realbox{running on real robots}, poses estimated and then shown in the simulator. Figure \ref{fig:kalman} is replicated with different metrics and visualization of the distance to belief for each point on the figure. The planning quality map (Figure \ref{fig:heatmaps}) is also reproduced, with control over the parameters of the density estimation. Figure \ref{fig:website} shows some of the tools available on the website.

\mysection{1}{Calculation of $D_{belief}$}
\label{ap:calculation_d_belief}

The \textit{distance to belief} measures the discrepancy between nominal trajectories within the in-domain environment and out-of-domain trajectories in the corrupted environment, hence modeling the impact of a change in configuration $\Delta E$. Formally, let us define a function $F_\theta: \cA\times\cP \mapsto \cP$ corresponding to the forward step of the environment parametrized by some physical parameters $\theta\in \Omega$. The function $F_\theta$ maps an action $a\in \cA$ and a current \textit{state} $\bmp_t\in\cP$ (position + velocity) to the future state $\bmp_{t+1}$. To simplify, $F_\theta$ models solely the dynamics of the robot, collisions are ignored to compute $D_\text{belief}$. 

\begin{figure*}
    \centering
    \includegraphics[width=\textwidth]{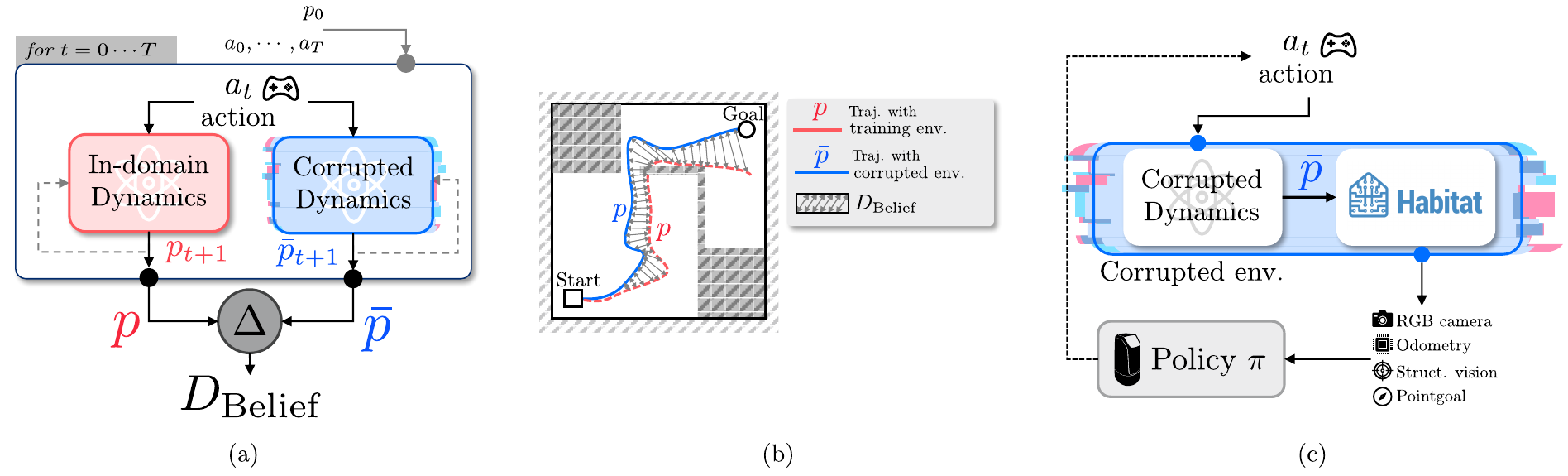}
    \caption{\label{fig:kalman_metric} \textbf{Comparing Prediction vs. Correction} -- steps in an end-to-end dynamic agent is achieved by testing the policy in an \textit{corrupted environment} where one of the step is made less accurate. Since changes in environment parameters $\Delta E$ are not comparable, we rely on the proposed \textit{distance to belief} to measure the impact of a change on the agent trajectory. (a) To compute this metric, we simulate trajectories generated by two different dynamical systems albeit from the same sequence of actions. (b) The distance to belief corresponds to the distance between the resulting trajectories without taking collisions into account. (c) While $D_belief$ is calculated by running an agent in the corrupted environment with the same actions as the agent had done in-domain, of course the actual success rate of the agent in the corrupted environment is calculated by letting the agent take its own decisions.} 
\end{figure*}

The \textit{distance to belief} is computed on a fixed set of $k\in\llbracket 1, K\rrbracket$ action sequences and an initial state $\{\bmp_0, a_0, \ldots, a_T\}_k$ where $T$ is the length of the sequences. The metric is defined as follows:
\begin{equation}
    D_{\text{belief}}{=}\frac{1}{TK} \sum_{t, k} \|\bmp_{t} - \bar{\bmp}_t\| \text{ s.t. } \left\{\begin{array}{ll}
        \bmp_{t+1} &= F_\theta(\bmp_t, a_t) \\
        \bar{\bmp}_{t+1} &= F_{\theta'}(\bar{\bmp_t}, a_t) \\
        \bar{\bmp}_0 &= \bmp_0,
    \end{array}\right. 
\end{equation}
where $\theta'$ is the set of corrupted environment parameters. In simple words, the distance to belief is proportional to the area between the in-domain and out-of-domain trajectories, and is measured in meters, so $D_{\text{belief}}=0.25$ can be interpreted as \textit{the corrupted trajectory diverges from the in-domain by \SI{0.25}{\meter} in average} (see Figure \ref{fig:kalman_metric}). We compute $D_\text{belief}$ in Figure \ref{fig:kalman} using $K=1,000$ sequences on $T=15$ steps (corresponding to \SI{5}{\second}). The action sequences are collected by sampling navigation episodes from the train set of HM3D solved by the model being studied (\oursbox{D28-dynamics} for Figure \ref{fig:kalman} (left), and \instantbox{D28-instant} for Figure \ref{fig:kalman} (middle)). The distance to belief is actually independent from the policy, which is only used to collect meaningful action sequences corresponding to realistic movements. The calculation of $D_\text{belief}$ only depends on the physical parameters of the environment.

As a rule of thumb, $D_\text{belief}$ values above \SI{1.0}{\meter} can be considered as highly corrupted environment, and reasonable values (arguably comparable with sim2real distance between the robot dynamics and the simulated model) lie in $[0, 0.5]$. Our interactive website shows the relation between the corrupted trajectory and the distance to belief.

\mysection{2}{Details on Prediction vs. Correction}
\label{app:kalman_details}
\begin{figure*}
    \centering
    \begin{tikzpicture}
        \draw (0, 0) node[inner sep=0] {\includegraphics[width=0.49\linewidth]{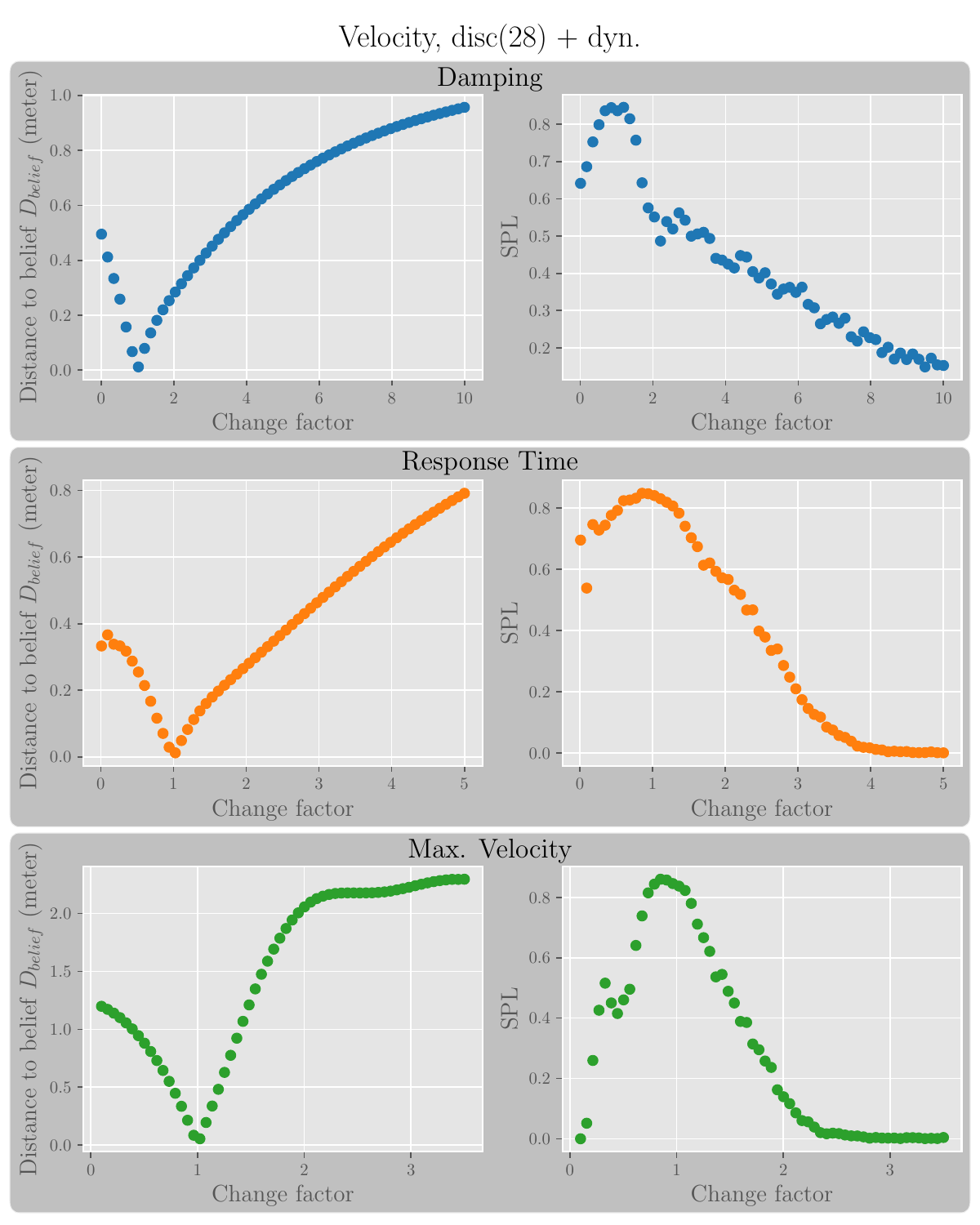}};
        \draw (8.5, 0) node[inner sep=0] {\includegraphics[width=0.49\linewidth]{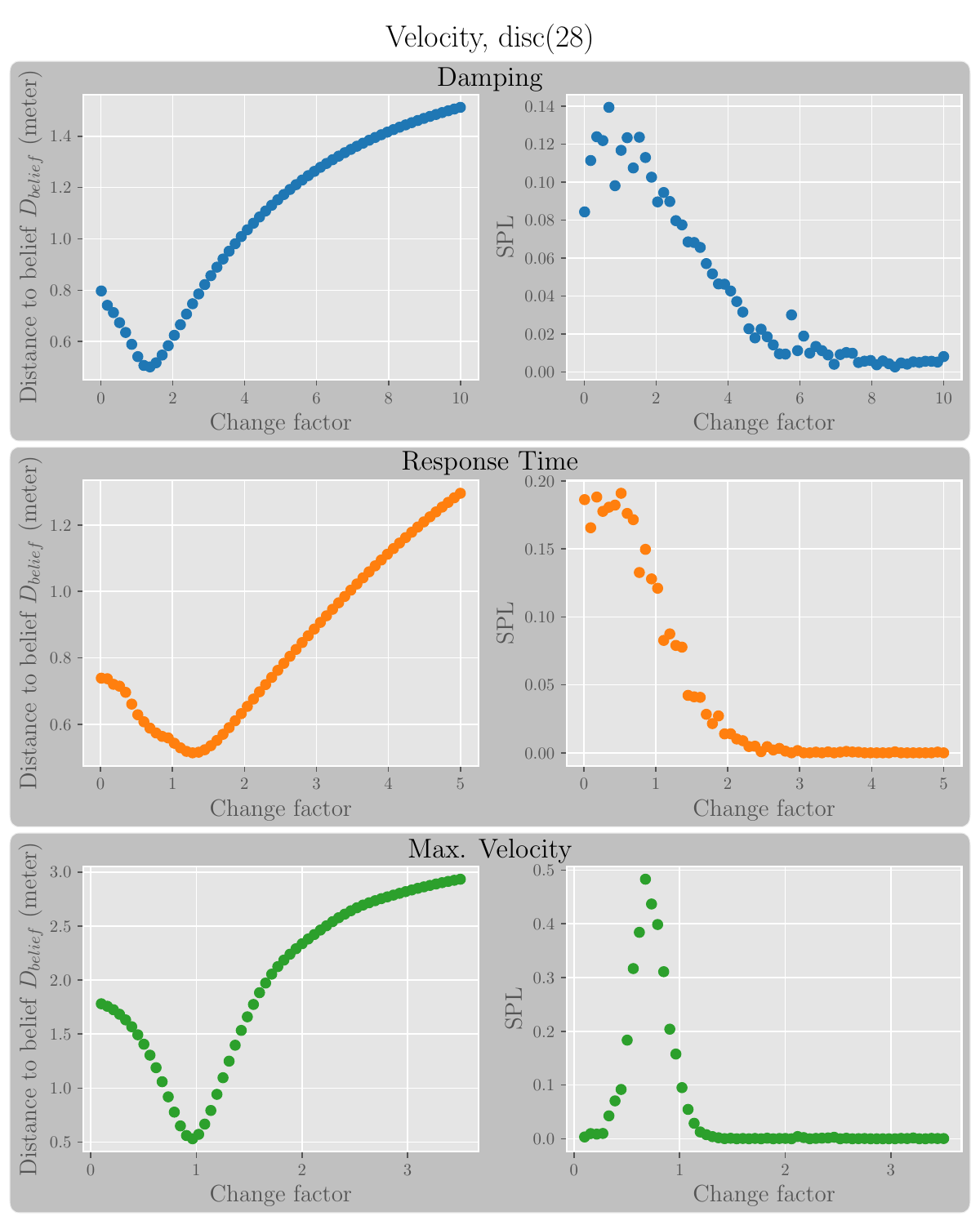}};
        \draw (0, 5) node[inner sep=0] {\footnotesize \oursbox{D28-dynamics/Table \ref{tab:variants}(c)}};     
        \draw (8.5, 5) node[inner sep=0] {\footnotesize \instantbox{D28-instant/Table \ref{tab:variants}(b)}};     
    \end{tikzpicture}
    \caption{\label{app:fig:kalman_details}\textbf{Detailed results Prediction vs. Correction} -- for \oursbox{D28-dynamics} (left) and \instantbox{D28-instant} (right). The figures shows, for each corruption type, the distance to belief and SPL score for different multiplicative factor $f$. We observe better robustness of the dynamic-aware agent against changes in the robot dynamics. Non-linear dependence between change factor and impact on the agent trajectory motivates the use of the distance to belief as a proxy to measure the effect of each corrupted environment.}
\end{figure*}

\subsection{\revision{Dynamical model}}
The dynamics of the real robot is modeled in the simulator using a second order dynamical model similar to \cite{bono2024learning}. Let $v(t), \omega(t)$ be the linear and angular velocity of the agent, and $a_v(t), a_\omega(t)$ be the linear and angular actions taken at time $t$. The dynamical model is 
\begin{equation}
    \label{eq:dyn}
    \begin{array}{ll}
        \ddot v(t) &= \frac{1}{\tau}\big(v(t) - a_v(t)\big) + \frac{2\gamma}{\tau}\dot v(t) \\
        \ddot \omega(t) &= \frac{1}{\tau}\big(\omega(t) - a_\omega(t)\big) + \frac{2\gamma}{\tau}\dot \omega(t), \\
    \end{array}
\end{equation}
where $\tau$ and $\gamma$ are the response time and damping factor. We apply different values depending on the motion direction (acceleration or braking) and type (linear or angular), resulting in four different constants for each parameter. We also apply saturation on acceleration (absent of this study) and on the velocity. \revision{This model is inserted between the policy and the simulator: the agent predicts a velocity command $(a_v, a_\omega)$ at a $3$Hz rate. This command is sent to the dynamical system which integrates \eqref{eq:dyn} (using a symplectic Euler scheme running at $30$Hz, the command is held constant during integration). The resulting position is then sent to the simulator (Habitat) which directly teleports the agent. We discretize the action space $[0, v_\text{max}]{\times}[0, \omega_\text{max}]$ into 28 discrete actions where $a_v \in \left\{0, \frac{v_\text{max}}{3}, \frac{2v_\text{max}}{3}, v_\text{max}\right\}$ and $a_\omega \in \left\{-\omega_\text{max}, -\frac{2\omega_\text{max}}{3}, -\frac{\omega_\text{max}}{3}, 0, \frac{\omega_\text{max}}{3}, \frac{2\omega_\text{max}}{3}, \omega_\text{max} \right\}$. We used the same numerical values as in \cite{bono2024learning}.}

Note that modification of the maximum velocity changes not only the saturation value, but also the distribution of discrete action. In other words, the 28 discrete actions are always scaled to fit the range of possible velocities. The dynamical model runs $10\times$ faster than the policy to prevent aliasing effects.

\subsection{\revision{Corrupted environments}}
Corrupted environments are generated by multiplying one of the physical parameters by a constant \textit{change factor} $f$, while leaving the other parameters untouched. Such a change results in a drop of performance on one hand, and an increase of the distance to belief on the other. As mentioned in the main paper, the factor $f$ causes a change in environment parameters $\Delta E$, which has different interpretations depending on the physical quantity and its impact on the dynamics. We unwrap Figure \ref{fig:kalman} and exposed the change factor in Figure \ref{app:fig:kalman_details}. In particular, we show the relationship between the change factor and the distance to belief, and its impact on SPL. We manually define suitable ranges for each corruption type (damping, max. velocity and response time) up to 0\% SR, and evaluate the agents (b) and (c) from Table \ref{tab:variants} on \epbox{HM3D/250} using linearly sampled change factors within the range.

Figure \ref{app:fig:kalman_details} shows that a fixed value of change factor $f$ can correspond to very different values of distance to belief depending on the corrupted parameter, which motivates the use of a proxy metric such as the distance of belief. We also observe steeper curves for SPL when testing the \instantbox{D28-instant} variant compared to \oursbox{D28-dynamics}, which confirms that training dynamic-aware agent allows the adaptation to different dynamic unseen during training.

\mysection{3}{Probing future pose: variants}
\label{sec:probingdynamics}
Our goal is to probe the existence of a plan in the latent state of the navigation agent. To do so, we collected a dataset of 500,000 navigation episodes generated from a trained \oursbox{D28-dynamics} agent and stored the latent state, action and path $(a_t,\bmp_t, \bmh_t)$. We split this dataset in proper train/validation/testing sets (80\%, 10\%, 10\%). During training, a random time instant $t$ is sampled in the episode, and the probing network is supervised to predict the future positions $\bmp_{t+1}, ..., \bmp_{t+H}$ from the first latent state $\bmh_t$. Future latent states $\bmh_{t+i}$ are not provided to the probing network, which can not use new observations to improve the predicted path. The probing network is trained to minimize
\begin{multline}
    \mathcal{L}_{\text{probing}} = \sum_{i=1}^H \overbrace{\left\|\begin{bmatrix} x_{t+i} \\ y_{t+i} \end{bmatrix} - \begin{bmatrix} \hat{x}_{t+i} \\ \hat{y}_{t+i} \end{bmatrix} \right\|^2_2}^{\text{pos. loss}} \\ +   \underbrace{\left|\begin{bmatrix} \cos \theta_{t+i} \\ \sin \theta_{t+i} \end{bmatrix} - \begin{bmatrix} \cos\hat{\theta}_{t+i} \\ \sin\hat{\theta}_{t+i} \end{bmatrix} \right|}_{\text{rot. loss}}
\end{multline}
assuming $\bmp_t = \begin{bmatrix} x_t & y_t & \theta_t\end{bmatrix}$ and $\hat{\bmp}_t$ being the predicted pose. We used the Adam optimizer (learning rate ${=}10^{-4}$) with a batch size of 64, a prediction horizon $H=20$ and performed 100,000 gradient updates. We tested different architectures of the probing network:
\begin{description}[labelindent=0mm,leftmargin=0mm,] 
    \item[Linear] uses a straightforward linear layer per time step to predict the future position from the initial latent state.
    \begin{align}    
    \forall i, \hat{\bmp}_{t+i} = \text{Linear}_i(\bmh_{t}), 
    \end{align}

    \item[Linear + non-linear(action,goal)] exploits a non-linear embedding of the previous action and the goal direction (given in polar coordinates with respect to the episode start).
    \begin{align}    
    \forall i, \hat{\bmp}_{t+i} = \text{Linear}_i \big( \bmh_{t}, \text{MLP}(\bma_{t+i-1}, \bmg)\big)
    \end{align}
    
    \item[GRU-agent] where we use the latent dynamics of the trained agent itself for prediction. Recall that the hidden state is updated by a GRU, cf. eq. (\ref{eq:gru}), which we reproduce here in a simplified notation without gates and only a single layer,
    \begin{equation}
        \mathbf{h}_t = \sigma(\mathbf{W} \mathbf{h}_{t-1} + \mathbf{V} \bmo_t),
    \end{equation}
    where $\mathbf{o}_t$ is a concatenation of all observation features, $\mathbf{W}$ is the matrix modeling latent dynamics, $\mathbf{V}$ projecting observations into the latent space, and $\sigma$ an activation function. Yet, since future observations $\bmo_{t+i}$ are not available during probing, we replace them by a transformation of the previous latent state performed by an MLP $\psi_\theta$ (3 layers, 1024 units, TanH activated) which compensates the absence of observations. 
    \begin{align}
    \begin{array}{ll}\mathbf{h}'_{t+i} = &\sigma\big(\mathbf{W} \mathbf{h}_{t+i-1} + \psi_\theta(\mathbf{h}_{t+i-1})\big) \\
    \hat{\bmp}_{t+\tau} = &\phi(\mathbf{h}'_{t+\tau}).
    \end{array}
    \end{align}    
\end{description}
All variants use a linear projection layer $\phi$ to map the latent space to the predicted position $\hat{\bmp}_{t}$.

\mysection{4}{Zeroing the hidden state of the agent}
\label{sec:zeroing}

In Table 4 of the main paper we described results ablating the memory of the agent, ie. setting the hidden GRU state $\mathbf{h}_t$ to zero. We argue that zeroing $\mathbf{h}_t$ only makes sense if at the same time we reset the episode specific coordinate frame of the agent, which defines the static goal vector $\mathbf{g}_0$ and the pose inputs $\hat{\mathbf{p}}^r_t$ and 
$\hat{\mathbf{p}}^a_t$. This is motivated by the fact, that the \textit{PointGoal} task requires the agent to understand where it is with respect to the goal, ie. it needs to have an understanding of the (unobserved!) dynamic goal vector $\mathbf{g}_t$, which is defined in its own egocentric frame. It can only do this by using the static goal vector $\mathbf{g}_0$ defined w.r.t. the episode start, and localization information, provided by $\hat{\mathbf{p}}^r_t$ and $\hat{\mathbf{p}}^a_t$, also in the same frame. The calculation can be done by a simple rigid transform, but the noise of these inputs will lead to a noisy signal. This noise can be filtered, for which the hidden memory of the agent is likely used. For this reason zeroing memory without resetting the episode centric coordinate can potentially lead to very undesirable effects.

\mysection{5}{Details on the planning heatmap}
\label{sec:heatmap}
In order to gain deeper insights into the specific areas where our robot encounters difficulties in navigation, we generated heatmaps that visually highlight challenging locations within our environment. This allows us to focus our efforts on improving the robot's navigation in these identified areas.
Below is the detailed description of the heatmap generation. 

Let \( \bmp_t = \begin{bmatrix} x_t & y_t & v_t & \omega_t \end{bmatrix}^T \) represents the state of an agent, where \( x_t, y_t \) are its 2D coordinates, \( v_t \) is the linear velocity, and \( \omega_t \) is the angular velocity. Given a goal \( \bmg \), we define \( \cT(\bmp_t, \bmg) \) as the \textit{time to goal}, which is the travel time to reach the goal. This value is computed by solving the Eikonal equation with fast marching, assuming the agent navigates at full speed and slows down near the walls. For each navigable point on the grid, the velocity is computed as $v(x, y) = V \times d(x, y) / K$ where $V$ is the max velocity of the agent, $d$ is the distance to nearest wall and $K=0.5$ a weighting coefficient. 

We introduce a cost function $\cC(\bmp_t, a)$ at state $\bmp_t$, taking an action $a$ as:
\begin{multline}
\cC(\bmp_t, a) = \overbrace{10\times\mathcal{T}(\bmp_{t+1}, \bmg)}^{\text{pos}} \\
+ \overbrace{0.1 \times \tan \left(\frac{\nabla_x \mathcal{T}(\bmp_{t+1}, \bmg)}{-\nabla_y \mathcal{T}(\bmp_{t+1}, \bmg)}\right)}^{\text{angle}} \\
+ \overbrace{\big(v_{t+1} - \beta\mathcal{T}(\bmp_{t+1}, \bmg)\big)}^{\text{slow down near goal}} \\ 
+ \overbrace{10^{-3}\times|\omega_t|}^{\text{rotation speed}}
+ \overbrace{10^{3}\times \mathcal{P}(\bmp_{t+1})}^{\text{collision}} 
\end{multline}

\medskip
\noindent where \( \bmp_{t+1} = \cD(\bmp_t, a) \) is the next state of the agent taking the action $a$ given by the dynamical model \( \mathcal{D}\), \( \cP(\bmp_t) \) the collision indicator, equal to $1$ if the agent collides and $0$ otherwise, and $\beta$ represents the braking strength, or how rapidly the agent can decelerate.

\medskip
\noindent Each term in the cost function has a specific purpose:
\begin{itemize}
\item \textbf{Position cost}: \( 10 \times \mathcal{T}(p_{t+1}, g) \) based on the time estimated to reach the goal.

\item \textbf{Angle alignment}: \( 0.1 \times \tan \left(\frac{\nabla_x \mathcal{T}(p_{t+1}, g)}{-\nabla_y \mathcal{T}(p_{t+1}, g)}\right) \) encourages alignment with the goal direction.

\item\textbf{Slowing near goal}: \( v_{t+1} - \beta \mathcal{T}(p_{t+1}, g) \) slows the agent down as it approaches the goal.

\item\textbf{Rotation speed cost}: \( 10^{-3} \times |w_t| \) discourages high angular velocities.

\item\textbf{Collision penalty}: \( 10^{3} \times \mathcal{P}(p_{t+1}) \) a large penalty applied if the agent collides.
\end{itemize}
This cost function is designed to balance reaching the goal quickly, maintaining alignment, slowing down near the goal, and avoiding high rotation speeds and collisions.  Then we can replay the recorded trajectory, knowing the taken action at every position, we calculate the following metric $M(t) =\cC(\bmp_{t+1}, a_{t+1}) - \cC(\bmp_t, a_t)$. To create a smooth, continuous heatmap, we apply a Gaussian kernel at each position \( (x_t, y_t) \), with a mean \( \mu = M(t) \) and standard deviation \( \sigma = 0.5 \). This results in a heatmap that provides a clear spatial representation of the robot's performance across the environment.

\mysection{6}{Evaluation in the training domain}
\label{sec:evaltraindomain}
We evaluate the three agents of Table \ref{tab:variants} also in a third setting: 
\tcbox[on line,colframe=white,boxsep=0pt,left=1pt,right=1pt,top=1pt,bottom=1pt,colback=blue!20]{\textbf{`` Simulation (train domain)''}} evaluates them in simulation w/o motion model, ie. with instantaneous velocity changes and constant velocities between time steps, and with their respective action spaces. This evaluates the difficulty of the training task and does not provide indications on  performance in a real environment.

\begin{table}[t]{
    \small 
    \centering
    \setlength{\tabcolsep}{1pt}
    \begin{tabularx}{\linewidth}{LPPPNNN}
        \specialrule{1pt}{0pt}{0pt}
        \rowcolor{TableGray2}
        \textbf{Method} &
        \multicolumn{3}{c}{\cellcolor{blue!50}\shortstack{Sim(train) \\ \epbox{HM3D/2.5k}}} &
        \multicolumn{3}{c}{\cellcolor{orange!60}\shortstack{Sim(+dyn) \\ \epbox{HM3D/2.5k}}} 
        \\
        \rowcolor{TableGray2}
        \textbf{}&
        {\cellcolor{blue!50}\bf  SR\%} & {\cellcolor{blue!50}\bf  SPL\%} & {\cellcolor{blue!50}\bf SCT\%} &
        {\cellcolor{orange!60}\bf  SR\%} & {\cellcolor{orange!60}\bf  SPL\%} & {\cellcolor{orange!60}\bf SCT\%} 
        \\
        \specialrule{1pt}{0pt}{0pt}
        \cellcolor{agentd4}
        \textbf{(a) D4} &
        {91.6} & {76.4} & {20.4} &
        {29.1} & {18.1} & {2.0}	 
        \\
        \cellcolor{agentinstant}
        \textbf{(b) D28-instant}  &
        {98.3} & {82.4} & {66.5} &
        {27.6} & {11.6} & {5.0}	
        \\
        \cellcolor{agentours}
        \textbf{(c) D28-dynamics}  &
        {97.6} & {82.3} & {52.2} &  %
        {97.6} & {82.3} & {52.2}
        \\
        \specialrule{1pt}{0pt}{0pt}
    \end{tabularx}
    }
    \caption{\label{tab:difficulty}\textbf{Evaluation in the training domain}. \trainbox{(Left)} the agent is evaluated in the same action space and dynamics used for training. This evaluates the difficulty of the task, and not the transfer to the real physical robot. \simbox{(Right)} the agent is evaluated in simulation with a dynamical model --- reproduced from Table 1 of the main paper.}
    \vspace*{-1mm}
\end{table}

\mysection{7}{Evidence of Tunnel vision}
\label{sec:tunnelvision}

We found some evidence of ``tunnel vision'', by which we mean that the agent attempts strategies, which a human could 
easily discard even without having access to a map. This is not necessarily a problem for successful completion of the episodes, as the agent detects blockings and searches for alternatives, eventually finding the goal. However, it is not efficient, and translates into lower then optimal SPL measures.

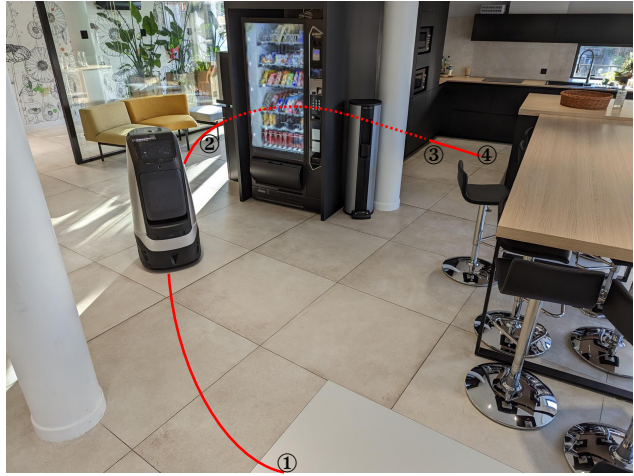
\begin{figure}[t] \centering
\begin{tikzpicture}
        \draw (0, 0) node[anchor=west,inner sep=0] {
        \includegraphics[width=\linewidth]{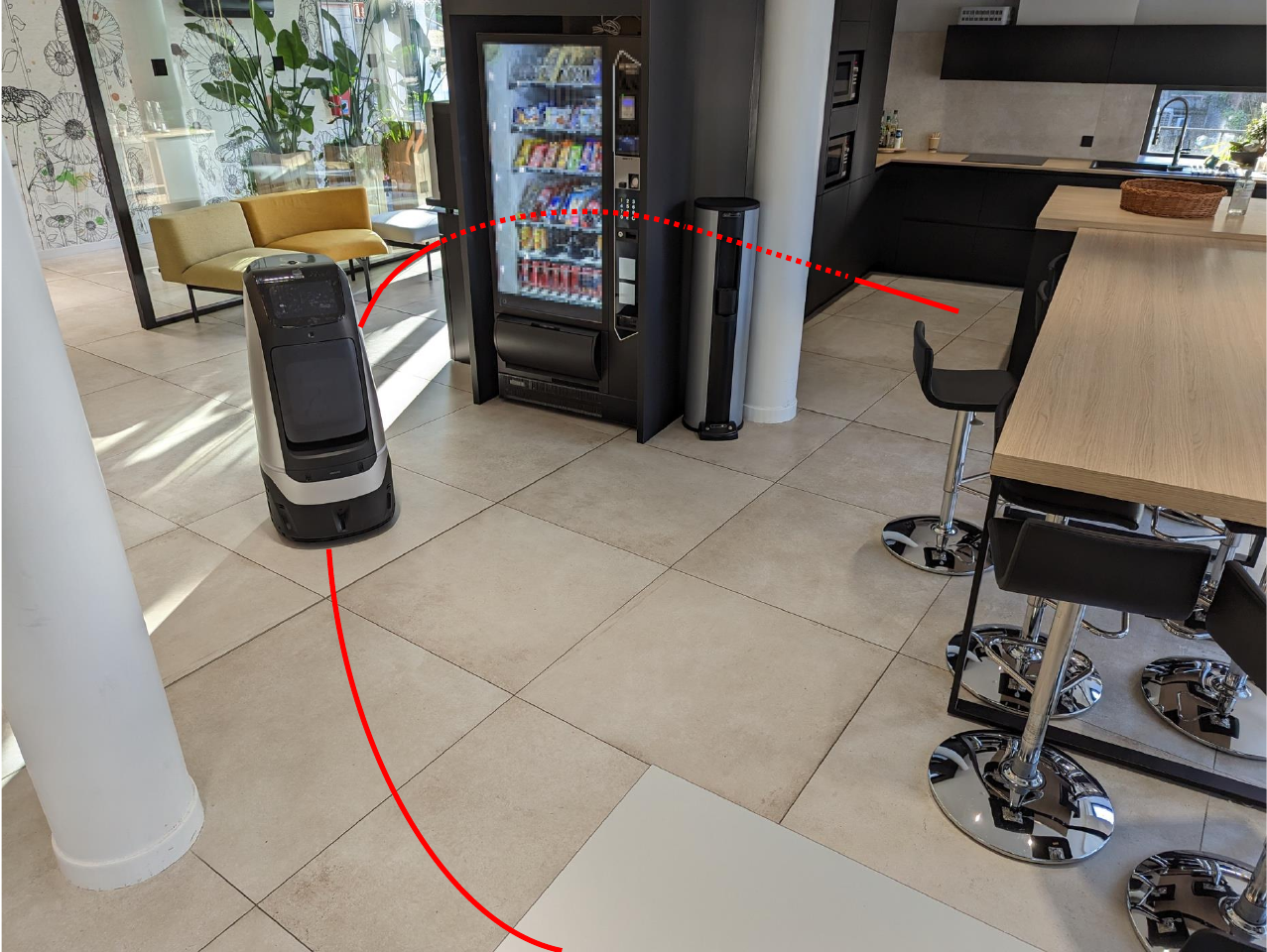}};      
        \draw (3.6, -3) node[anchor=west,inner sep=0] {\ding{172}};
        \draw (2.57, 1.25) node[anchor=west,inner sep=0] {\ding{173}};
        \draw (5.55, 1.1) node[anchor=west,inner sep=0] {\ding{174}};
        \draw (6.26, 1.1) node[anchor=west,inner sep=0] {\ding{175}};
\end{tikzpicture}
\caption{\label{fig:tunnelvision}\textbf{Tunnel vision:} the agent attempts to navigate along the red trajectory, although it is clearly visible that it is blocked, although the dotted part is occluded.}
\end{figure}

An example is seen in Figure \ref{fig:tunnelvision}. In this episode, starting at position \ding{172} and aiming for the goal position at \ding{175}, the agent tries to pass through the path indicated by the red trajectory, doing a turn into the area indicated by \ding{174} although it is clearly visible (from position \ding{172} already), that there is no path possible between \ding{173} and \ding{174}. Although the dotted part is occluded, a human would be able to estimate that it is blocked.

\mysection{8}{Details on visual localization}
\label{sec:vloc}

As an alternative to Adaptive Monte-Carlo Localization (AMCL)\cite{thrun2005probabilistic}, we experimented with a custom visual localization system, cf. Table 5 in the main paper. Here we provide more details on the setup.

For the pre-mapping part, we follow a procedure similar to the one describe in~\cite{Lee_2021_CVPR}: a dedicated robot is driven through the environment capturing synchronized 3D LIDARs, RGB cameras and odometry data. Both standard \textit{Simultaneous Localization And Mapping} (SLAM) using \textit{Iterative CLosest Point} (ICP) on the LIDAR point-clouds and \textit{Structure-from-Motion} (SfM) matching local features in RGB frames are used to recover the poses of all RGB frames relative to an absolute, unified, coordinate system.
An elastic-search database stores the 12k RGB frames, associated with a global descriptor, and local descriptors of keypoints computed by fast-R2D2~\cite{R2D2_2019}.

The navigating agent can then query the visual localization system by sending an image captured by its own RGB sensor and its last pose estimate, which locally combines the results of the last visual localization query and odometry. First, as described in~\cite{kapture2020}, the global descriptor for the image is used to quickly retrieve a set of nearest neighbors from the database. Then local R2D2 key-points in the image are matched against the ones of the retrieved neighbors and from this relative poses estimates and the absolute camera poses of the neighbors, a consensus is established for the absolute pose of the camera of the agent, which is sent back to the agent, where it is fused into its own localization optimization graph (with previous loc and odometry).

\mysection{9}{\revision{Architecture and training details}}
\label{sec:model-details}
\revision{\myparagraph{Training} all variants are trained on a single Nvidia A100-80Gb GPU for 500 million steps (24 environments in parallel, 192 steps per rollouts, 4 epochs per rollouts split in 2 batches). The agent is trained using PPO  with the following reward function: $r=R\mathbb{I}_\text{success} - \Delta_t^\text{Geo} - \lambda - C\mathbb{I}_\text{collision}$ where $R=2.5$, $\Delta_t^\text{Geo}$ is the gain in distance to the goal, $\lambda=0.01$ to prevent the agent from stalling and $C=0.1$ to penalize collisions.} 

\revision{\myparagraph{Agent architecture} the RGB encoder is a ResNet18 with 64 base planes and 4 layers of 2 basic blocks each, and the scan encoder is a 1d-CNN composed of 3 Conv-ReLU-MaxPool blocks with 64, 128 and 256 channels, kernels of sizes 7, 3 and 3, circular padding of sizes 3, 1 and 1, and pooling windows of widths 3, 3 and 5. The encoder ends with a linear layer to flatten the representation to an embedding size of 512. Odometry, localization, and goal are encoded using MLPs with a single hidden unit of size 1024 and an output size of 64 (ReLU activation). Angles are transformed into cos/sin representation before being fed to the network. The action encoder is a discrete set of 29 embeddings of size 32, and the state encoder is a 2-layer GRU with hidden state of size 1024. The policy is a fully linear layer applied on the hidden state of the last layer of GRU.}

\end{document}